%% file: main.tex
\documentclass[a4paper,conference]{IEEEtran}
\usepackage{cite}

\ifCLASSINFOpdf
  \usepackage[pdftex]{graphicx}
\else
\fi
\usepackage{amsmath}
\ifCLASSOPTIONcompsoc
 \usepackage[caption=false,font=normalsize,labelfont=sf,textfont=sf]{subfig}
\else
 \usepackage[caption=false,font=footnotesize]{subfig}
\fi
\usepackage{amssymb}     
\usepackage{tikz}
\usepackage{xcolor}
\usepackage{multirow}
\usepackage{eucal}
\usepackage{makecell} 
\usepackage{booktabs} 
\usepackage{siunitx}
\usepackage{microtype}

\DeclareSIUnit\px{px}

\renewcommand\IEEEkeywordsname{Keywords}

\def\secref#1{Sec.~\ref{#1}}
\def\figref#1{Fig.~\ref{#1}}
\def\tabref#1{Tab.~\ref{#1}}
\def\eqref#1{Eq.~(\ref{#1})}

\renewcommand{\d}[1]{{\mbox{\boldmath$#1$}}}
\newcommand{\m}[1]{{\mbox{{\fontencoding{T1}\sffamily\slshape{#1\/}}}}}

\newcommand{\ical}[1]{{\mbox{\usefont{OT1}{pzc}{m}{it}{#1}}}}

\begin{document}


\title{Temporal Prediction and Evaluation of Brassica Growth in the Field using Conditional\\ Generative Adversarial Networks}

\author{
    \IEEEauthorblockN{
        Lukas Drees\IEEEauthorrefmark{1},
        Laura Verena Junker-Frohn\IEEEauthorrefmark{2}, 
        Jana Kierdorf\IEEEauthorrefmark{1}, 
        Ribana Roscher\IEEEauthorrefmark{1}
    }\\
    
    \IEEEauthorblockA{\IEEEauthorrefmark{1} IGG, Remote Sensing, University of Bonn, Germany; \{ldrees, jkierdorf, ribana.roscher\}@uni-bonn.de}
    \IEEEauthorblockA{\IEEEauthorrefmark{2} IBG-2, Plant Sciences, Forschungszentrum Jülich GmbH, Germany; l.junker-frohn@fz-juelich.de}}






\maketitle

\begin{abstract}
Farmers frequently assess plant growth and performance as basis for making decisions when to take action in the field, such as fertilization, weed control, or harvesting.
The prediction of plant growth is a major challenge, as it is affected by numerous and highly variable environmental factors.
This paper proposes a novel monitoring approach that comprises high-throughput imaging sensor measurements and their automatic analysis to predict future plant growth.
Our approach's core is a novel machine learning-based generative growth model based on conditional generative adversarial networks, which is able to predict the future appearance of individual plants.
In experiments with RGB time series images of laboratory-grown \textit{Arabidopsis~thaliana} and field-grown cauliflower plants, we show that our approach produces realistic, reliable, and reasonable images of future growth stages.
The automatic interpretation of the generated images through neural network-based instance segmentation allows the derivation of various phenotypic traits that describe plant growth.
\end{abstract}
\begin{IEEEkeywords}
generative adversarial networks, agriculture, cauliflower, prediction, plant growth
\end{IEEEkeywords}

\section{Introduction} \label{sec:intro}
Digital solutions contribute to increase agricultural productivity and improve yield security, especially during changing climate conditions \cite{fischer2009world,gebbers2010precision,kitzes2008shrink,tyagi2016towards}.
In recent years, machine learning has become increasingly applicable in the agricultural sector.
Especially deep learning methods help to analyze crop phenotypes at an early stage and to detect plant diseases and weeds to enable a targeted removal without pesticides \cite{kamilaris2018deep,zhu2018deep}.
In addition to classification and regression tasks, like the distinction between crops and weeds and the determination of biomass from images, temporal prediction plays an increasingly important role.
A prediction of growth stages at an early date, for example, allows planning security but also enables the farmer to adapt the planned setup objectively through comprehensive spatial and temporal information \cite{chlingaryan2018machine,kuwata2015estimating,you2017deep}.

A novel way to predict and analyze the future appearance of a scene or individual objects like plants is the use of generative adversarial networks (GANs, \cite{goodfellow2014generative}).
GANs consist of two neural networks, the generator, and the discriminator. 
While the generator tries to create realistic images, the discriminator tries to distinguish these generated images from real ones \cite{ledig2017photo,radford2015unsupervised}.
Both networks are optimized together with the result that a powerful generator is learned that can generate images with respect to specifically targeted properties like the growth stage of plants.
In this case, contrary to classical growth models, generated images of future plants have two advantages:
First, numerous parameters of interest can be derived by statistical analysis or further application of machine learning image interpretation methods. 
This is an advantage over many classical approaches, which often estimate only individual parameters.
Second, images are directly assessible and increase the reliability of the results because they can be visually interpreted by farmers, which is in line with the goals of explainable machine learning \cite{roscher2020explain}.
A general advantage of GANs for temporal prediction is that they are trained using time series that can be acquired by regular measurements with satellites, UAVs, or ground robots and do not require time-consuming annotations of images.

\begin{figure*}[tb]
\centering
\captionsetup[subfigure]{labelformat=empty}
\subfloat[Ladybird on experimental site]{\includegraphics[height=5.0cm]{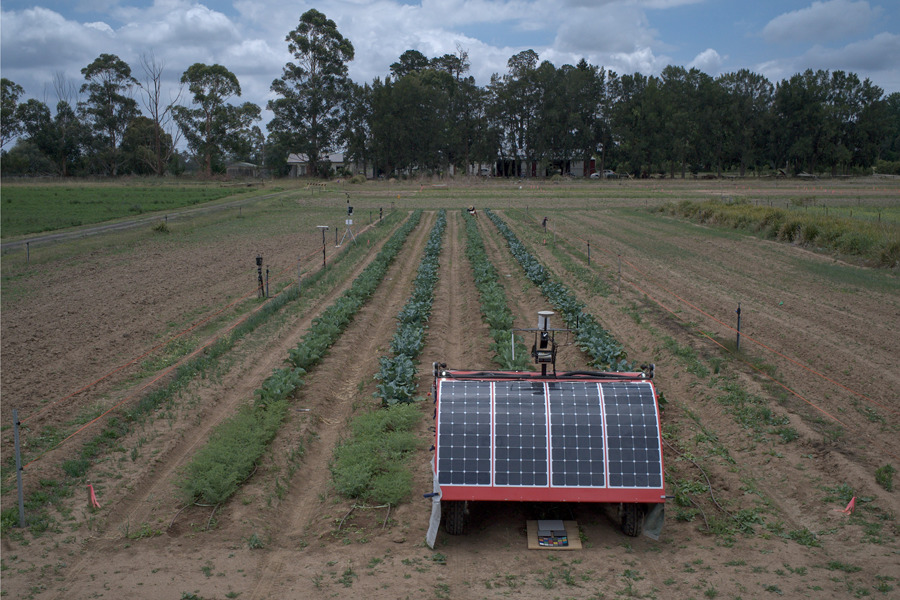}}
\hspace{1cm}
\subfloat[Schematic views of Ladybird]{\includegraphics[height=5.0cm]{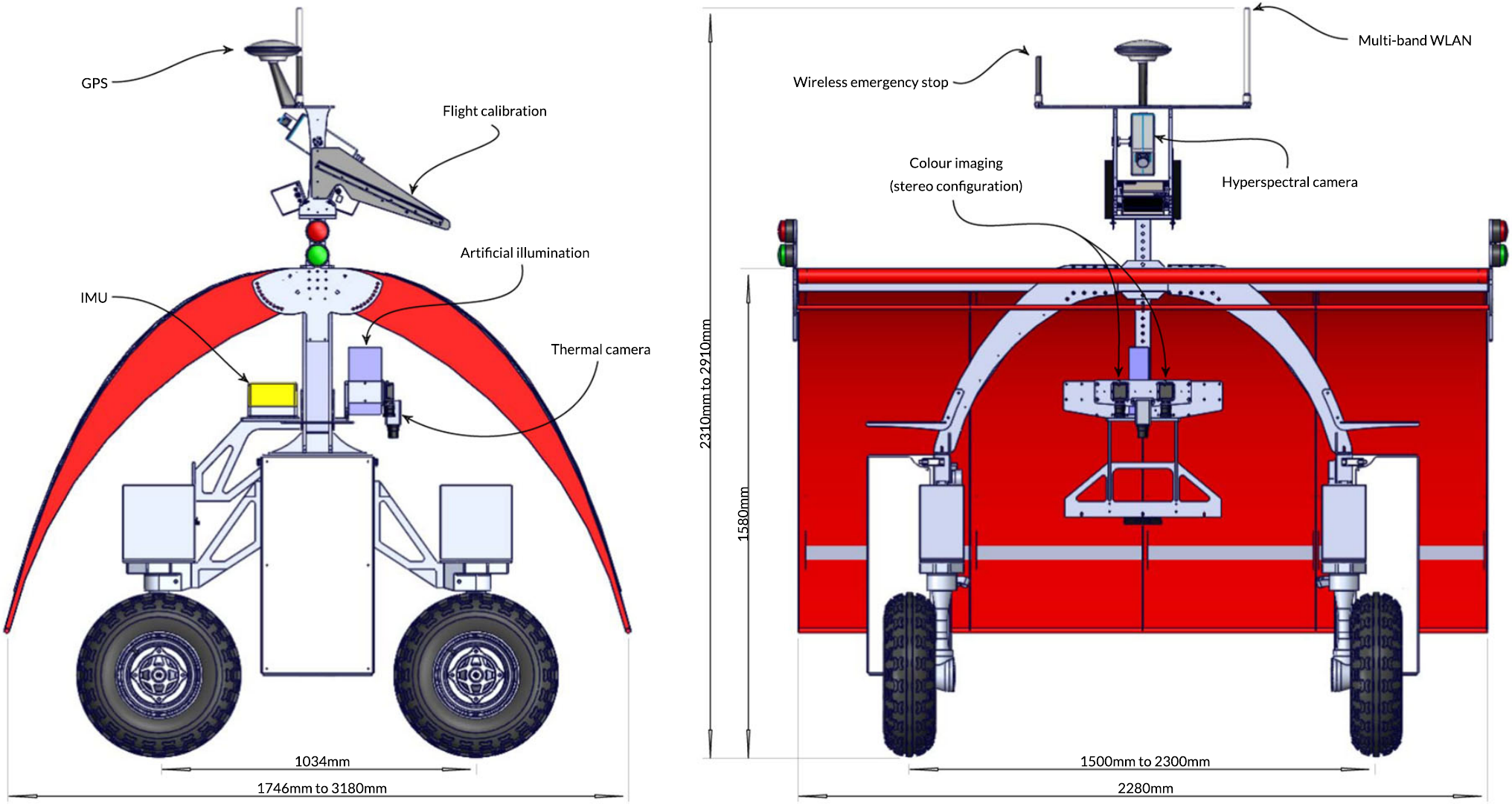}}
\caption{The Ladybird kinematic multi-sensor system was used for image acquisition. On the left, in action at the experimental site of the Lansdowne Farm in Cobbitty, near Sydney, Australia; on the right the schematic views with the sensors installed. All images are from \cite{bender2020high}.}
\label{fig:ladybird}
\end{figure*}

In this paper, we predict future growth stages of plants based on RGB-images acquired in the field using a conditional GAN (cGAN) \cite{isola2017image}.
Throughout this paper, we refer to the growth stage as the visible phenotype of the plant at different plant ages.
As condition for the generation of the images of future growth stages, we use the appearance of a plant at one date in the past.
The pipeline consists of (1) training a data-driven generative growth model using a cGAN, (2) temporal prediction of plants from earlier to later growth stages, and (3) a thorough qualitative and quantitative evaluation of the plant traits derived by instance segmentation and the generated images by Fréchet Inception Distance (\text{FID}, \cite{heusel2017gans}).

We show that the learned cGAN is highly capable of learning different data-driven generative growth models of \textit{A.~thaliana} and cauliflower and predicting realistic looking and reasonable images.
By realistic, we mean that the appearance of the generated plant images is not distinguishable from reference plant images at the same growth stage.
Reasonable means that plant traits derived from the generated images are in line with traits assessed of reference plants. 
Moreover, generated images serve as visual support for assessing the reliability of the estimation.
Our results are objectively evaluated by \text{FID} score between generated and real plant images. 
We further evaluate them using Mask~R-CNN \cite{wu2019detectron2} by performing instance segmentation on real and generated images.
For cauliflower, the projected leaf area derived from generated images revealed different growth patterns between plants exposed to differently fertilization and irrigation treatment, indicating the potential of the method to conclude from earlier plant growth on future plant growth under different environmental conditions.

\begin{figure*}[t]
	\centering
	\includegraphics[width=1\textwidth]{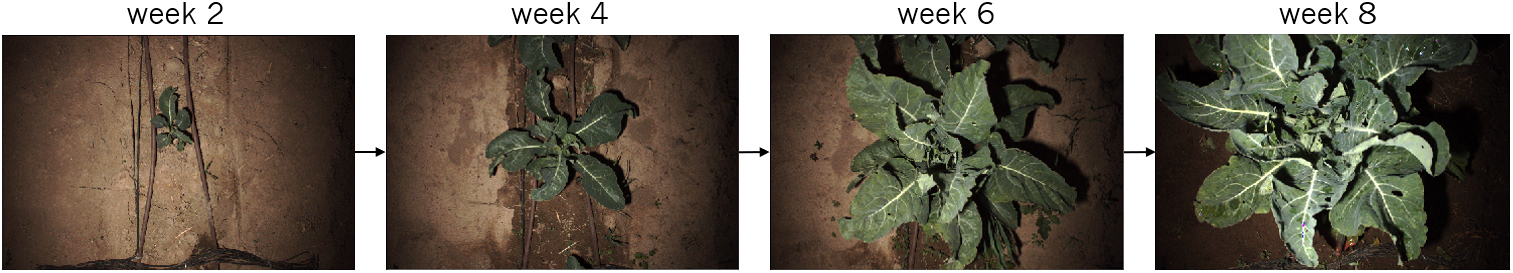}
	\caption{Aligned images of a cauliflower plant 2, 4, 6, and 8 weeks after planting. Please note, that the plant's brightness increases from week to week as it grows higher and is, therefore, closer to the artificial light source of the ground robot. Moreover, in later growth stages the plant goes beyond the size of the image and cannot be captured entirely.}
	\label{fig:data_weeks}
\end{figure*}

\subsection{Related Work}\label{sec:relwork}
\subsubsection{Conventional plant growth prediction}
Growth and phenological development of crops is commonly assessed using the BBCH scale that describes the growth stages from germination to senescence \cite{feller1995phanologische}. Plant growth is affected by abiotic factors such as temperature, precipitation and further climate conditions, and biotic factors like pests and pathogens \cite{pandey2017impact}. 
Many different models have been developed to predict crop plant growth based on the estimation of photosynthesis, temperature sums, or different climate conditions \cite{kage2001predicting,miller2001using,olesen2000simulation}. These conventional approaches are knowledge-based and estimate plant growth on external factors \cite{sihag2019review}. In contrast, the cGAN approach is based on the phenotype of plants, which integrates plants’ adjustments to the abiotic and biotic factors, which it was previously exposed to.

\subsubsection{Plant growth prediction with machine learning}
Machine learning methods are potential tools to solve the challenges in predicting plant growth.
For this purpose, recurrent neural networks (RNNs) and long short-term memory (LSTMs) have shown to be especially suitable due to their ability to process temporal information. 
Earlier approaches, as presented in \cite{park2005comparison}, have shown that they are applicable for predicting reasonable yield values from soil quality measurements or for modeling plant growth and selected parameters adequately, such as as the trunk diameter of tomato plants \cite{alhnaity2019using}. 
The similarity between both methods is the type of input data, where one-dimensional measurement values or parameters are used to train the artificial neural networks.
However, prediction also works for two-dimensional input, i.e. UAV images. 
This has the advantage that the plants are observed more comprehensively, and characteristics of all visible plant traits are gathered into the model \cite{johansen2019predicting,nevavuori2019crop}.

\subsubsection{Generating plant images}
For temporal analysis, generative models enable to predict the plant phenotype at a future point in time. 
In doing so, parameters such as diameter, biomass, or yield can be derived from the generated images.
There are three advantages in generating whole plants rather than predicting individual parameters.
The first advantage refers to the fact that reference data, which is generally needed to learn prediction models for individual parameters, is time- and cost-intensive to obtain.
GANs overcome the challenge, since they typically learn in an unsupervised way.
Not needing reference data is a major advantage from a practical perspective, as typically, reference data only exists for harvested fields, when the harvest is counted or weighed. 
For the implementation of our proposed method in agricultural practice, fields need to be monitored, e.g. by UAV images, a well-established and easily applicable technique \cite{watt2020phenotyping}.
In this context, the second advantage is that UAV imaging allows not only for a monitoring of plants at all growth stages, but also reveals spatial information and quantifies potential heterogeneity in the field.
Third, if images instead of parameters are predicted, this also includes prediction of developing plant organs such as flowers, as well as effects of abiotic and biotic stresses on plant morphology. Thereby, negative effects of stresses can be detected early \cite{foerster2019}.
This enables selective treatments of individual plants rather than treating whole fields.

\subsubsection{GANs in agriculture}
The generation of artificial images with GANs has recently found increased attention in agriculture and plant science.
The underlying challenge and task of GANs are to obtain a translation between two domains, A and B, where a so-called domain is a set of data samples such as images whose distribution is implicitly determined by the GAN.
The various GAN approaches presented in the literature differ in how domains A and B are chosen. 
In our work, we will refer to the early plant stage as the source distribution, denoted as domain A, and the advanced plant stage as the target distribution denoted as domain B, where both domains are included in the training dataset and images from domain B are generated when applying the trained method.

A commonly used type of GANs for agricultural applications are cycle-consistent generative adversarial networks (CycleGANs), e.g., for the detection and discrimination of plant diseases and the estimation of their future spread on its leaves \cite{foerster2019,li2018unsupervised,nazki2020unsupervised}.
Other applications include the translation of real images (domain A) into outputs (domain B) that directly contain interpretations of the data such as semantic segmentations \cite{barth2018improved}. 
Furthermore, outputs can also be products such as vegetation indices like the normalized difference vegetation indices \cite{suarez2019image}.
Image-to-image translation is also suitable for data augmentation and up-scaling of plant imagery, which produces new higher-resolution images from low-resolution ones and thus enables to analyze plant traits in a more detailed way \cite{dai2020crop,nazki2019image}.

CycleGANs are particularly suitable if necessary to translate in both directions, from domain A to domain B and vice versa.
Thereby, they do not require aligned image pairs, which means that for an image from domain A there does not have to exist a corresponding image of the same plant in domain B \cite{zhu2017unpaired}.
The ability to use non-aligned data is essential for many applications that have sufficient training data from both domains, e.g., leaves with and without disease, but which have only a few image pairs \cite{foerster2019}.

\subsubsection{Conditional GANs}
In agriculture, aligned temporal image pairs are becoming widely available due to geo-referenced orthophotos or by using kinematic multi-sensor systems, which help to position sensor imaging data, for example, by GPS.
In order to exploit this specific data characteristic, cGANs can be used to learn a powerful generator based on a given set of input and output pairs \cite{isola2017image}.
These networks show for various application areas that they can achieve good results in the field of domain adaptation \cite{hong2018conditional,lin2018conditional,mirza2014conditional}, but they have rarely been used in plant science so far.

Most machine learning methods and especially deep neural networks such as GANs generally require vast amounts of data to learn from \cite{sun2017revisiting}.
In case that a limited number of data is available, data augmentation can be used to stabilize the learning process of the model.
A few specific applications already successfully use plant data in cGANs, like artificial targeted plant generation, for the aim of data augmentation.
For this purpose, \cite{zhu2018data} use segmentation masks of plants on the input side of a cGAN to synthesize new real looking plant images on the output side. 

One advantage of cGANs is that a condition can be introduced which actively influences the appearance of the generated images.
In the simplest case, the condition is an image based on which a new image is to be generated. 
However, this approach is highly versatile, since the condition and the generated output do not have to have the same dimension.
Beside pairs of images of the same size, also other corresponding pairs of condition and outputs can be used, e.g., when generating images (output) from scalars or vectors (condition).
For instance, \cite{giuffrida2017arigan} generate images with plants of different sizes, where the number of leaves is introduced as a scalar condition attached to a noise vector.
The resulting generated images can be used for data augmentation to correct imbalances and to generate adequate samples in the training set.

To our knowledge, there is no related work yet on the utility of agricultural data pairs in cGANs where the domains differ in time, as is done in this work.
However, methodically related to our work is face aging, which involves predicting faces several years into the future using cGANs \cite{antipov2017face}. Here, cGANs help maintain the characteristics that form a person's identity in the future, just as it is crucial for plant traits.

\begin{figure*}[t]
	\centering
	\includegraphics[width=1\textwidth]{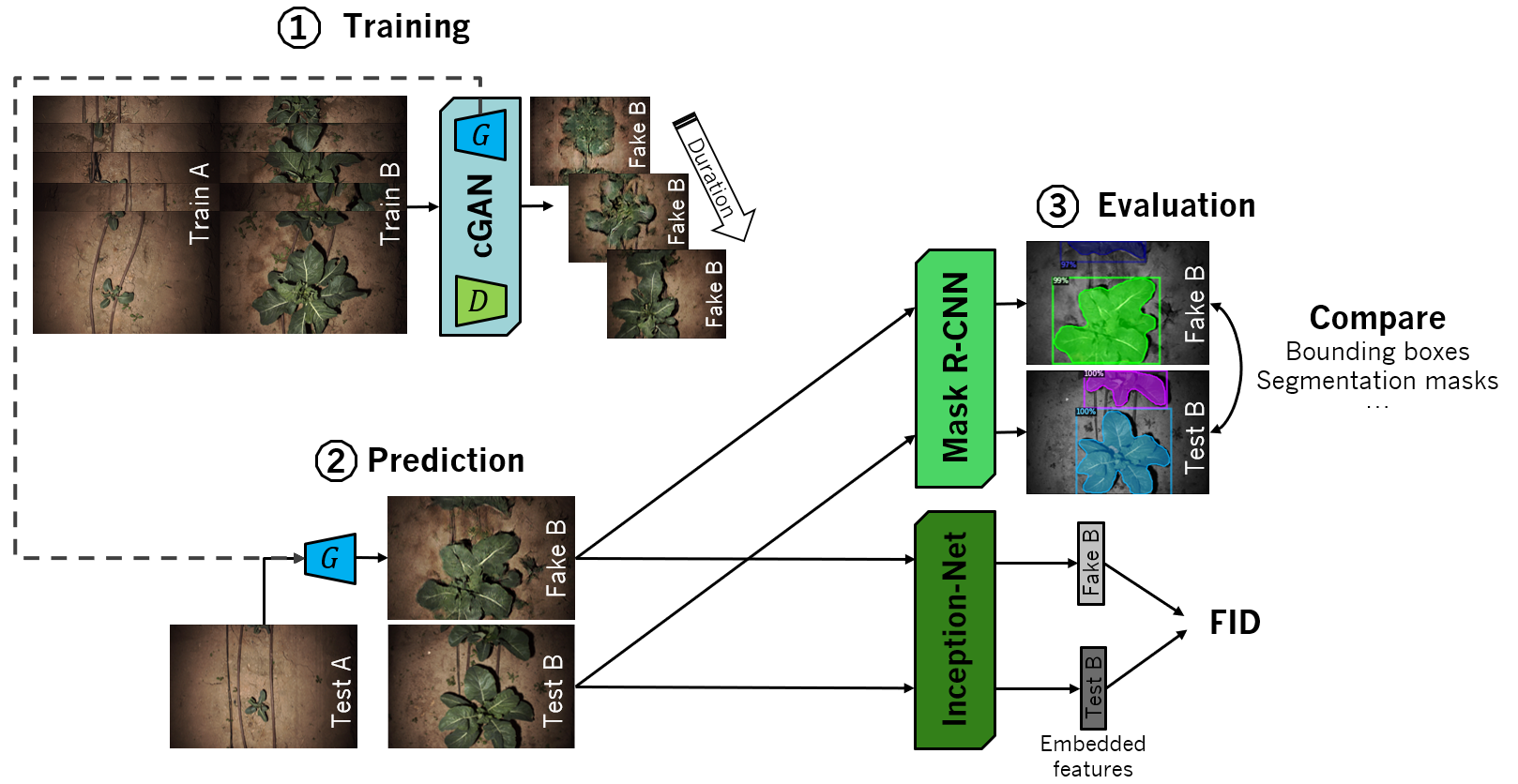}
	\caption{Processing pipeline: First, a conditional GAN is trained on training pairs of domains A and B. Through adversarial training of discriminator (\ical{D}) and generator (\ical{G}), the generated images become more realistic with each epoch. Second, the generator is used to generate predictions from the test-input. The third step is the evaluation, divided into instance segmentation using Mask~R-CNN and \text{FID} calculation. Instances are calculated on generated images as well as on reference images with the use of a Mask~R-CNN model. The comparison of the instance parameters (bounding box, area of the segmentation mask) allows a statement about the quality of the generator. \text{FID} score provide an additional objective measure about the quality of the generated plant images.}
	\label{fig:pipeline}
\end{figure*}

\subsubsection{GAN evaluation}
To ensure that the evaluation of GAN-generated images is not left to human subjective assessment, a wide range of evaluation methods has been developed \cite{borji2019pros}. 
This is particularly important because the loss curves of adversarial training and validation are more challenging to interpret than those of classical neural networks so that additional means of evaluation are needed \cite{goodfellow2014generative}.

The approaches discussed by \cite{borji2019pros} compare the real images and the generated images directly, e.g., using the L1-norm, or compare the real data distribution with the generated distribution, e.g., using the Wasserstein distance \cite{arjovsky2017wasserstein} or the Fréchet inception distance (\text{FID}, \cite{heusel2017gans}).
The FID score, which is also used in this work, is more robust to noise than other standard evaluation measures like the Inception Score \cite{salimans2016improved} and works well for the aspects discriminability, robustness, and computational efficiency. 
However, \text{FID} considers the image in its entirety, and the less important background has a particular influence on this metric.

Another possibility, which is used in this paper, is to compare parameters that are derived from real and generated images.
In this way, generated images can be analyzed and evaluated by their content, for example, by segmentation of whole plants instances or plant organs like leaves and fruits \cite{roscher2014Automated,zabawa2020counting}.
For this purpose, a wide variety of image-based analysis methods in the deep learning area can be used that work on plant detection in the field \cite{hamuda2016survey}.
Mask~R-CNN is the current state-of-the-art for calculating instance segmentation and is therefore used in this work \cite{he2017mask}.

\section{Material and Methods}\label{sec:mat_met}

\subsection{Data}\label{sec:data}
In the following, we use two different datasets. 
The Aberystwyth leaf evaluation dataset of \textit{A.~thaliana} plants \cite{bell_jonathan_2016_168158} was created in the laboratory and therefore has a high number of data, constant lighting, and no plant overlaps.
In comparison, the Brassica dataset \cite{bender2020high} of field-grown cauliflower plants reflects the typical challenges of real field measurements. 
The number of images is lower, the sensor calibration fails in one week, spatial alignment is less accurate, and crop growth depends on more external factors (climate conditions, soil quality, cultivation) than under laboratory conditions.

For our approach, we define image pairs of two domains A and B, where domain A is an image of a plant at an early growth phase and domain B is an image of a plant at a later growth phase. 
The terms training data and test data refer without domain specification always to aligned image pairs of domain A and B.
To indicate individual images, we refer to those that serve as the condition in the cGAN as input (train-input, test-input) and those with the same domain as the generated image as the reference (train-reference, test-reference). For a better reading, if not specified, the term reference refers to test-reference.

\subsubsection{Aberystwyth leaf evaluation dataset}
The Aberystwyth leaf evaluation dataset contains images of the plant \textit{Arabidopsis~thaliana}, which have been grown in a greenhouse \cite{bell_jonathan_2016_168158}.
A robot equipped with an RGB-camera acquired images of four trays with 20 plants each at regular intervals of 15 minutes during the day.
The plants were observed from day 21 after sowing to day 55 after sowing.
For the generation of image pairs, a spatial and temporal alignment is given because the camera and trays position does not change over the time series, and the sampling intervals are approximately constant.
For two trays, there are labels on leaf-level provided, which are used to train and test the Mask~R-CNN for evaluation.

\subsubsection{Brassica dataset}
The cauliflower imagery used in this study is taken from the Brassica project, which contains comprehensive multi-modal data from autonomous field observations and in situ measurements of crop sizes \cite{bender2020high}. The experimental site was in Australia about 70 km southwest of Sydney at Lansdown Farm, see \figref{fig:ladybird} left.
There, two beds of cauliflower (\textit{Brassica oleracea var. botrytis}) with about 140 plants each were imaged from sowing to harvest.
Both beds are divided into about 4 equally sized subareas of different irrigation and fertilization. In the following the subareas are indicated with \{\texttt{i+f+}, \texttt{i+f-}, \texttt{i-f+}, \texttt{i-f-}\}, where \texttt{i} denotes irrigation, \texttt{f} fertilization and \texttt{+}/\texttt{-} mean sufficient or insufficient conditions, respectively.
The imaging was performed weekly in approximately equidistant intervals over ten weeks from planting to harvest.
The measurement vehicle is a kinematic multi-sensor system called Ladybird, more precisely an autonomously driving ground robot equipped with localization (IMU, GPS), communication (WLAN), and camera sensors (RGB, hyperspectral, thermal), see \figref{fig:ladybird} right. 
This work focuses on the RGB images from two cameras taken in a stereo setting from above using an artificial light source while Ladybird is moving at a speed of 0.1 m/s.
The acquisition rate was 0.5 Hz in the first part of the growth phase (weeks 1-4) and 1 Hz in the second half (weeks 5-10) resulting in 4 resp. 8 images of the same plant in the maximum.
Images from the growth stage of week 1 to week 9 of cauliflower that are available for each bed are used, while in the growth stage of week ten, most cauliflowers extend beyond the edges of the image and are therefore not considered.

Due to Ladybird's localization sensors it is possible to assign geo-referencing to each acquired image.
This enables the monitoring of images of the same plant at different times.
\figref{fig:data_weeks} shows the RGB image of an identical plant over weeks 2, 4, 6, and 8 of the growing season.
One can see that from a certain size, the plant go beyond the size of the image and cannot be captured entirely, even if it grows in the image center.
It is also visible that the plant appears brighter with each week because it has grown closer to the robot's light. 
From the geo-referencing of the images, aligned image pairs are derived, i.e., images that have observed the same scene in at least two different growth stages.
We consider two images as showing the same scene at different points in time if the GNSS-derived coordinates of the image centers do not differ by more than $2cm$.
This corresponds to the accuracy level of the ground robot's geo-referencing. 
As the robot did not move in exactly the same line each week, about 50~\% of the images had to be excluded from the dataset.
The aligned image pairs serve as input for training and testing the cGAN.

\begin{figure}[t]
\captionsetup[subfigure]{labelformat=empty}
\centering
\subfloat[generated A]{\includegraphics[height=2.8cm]{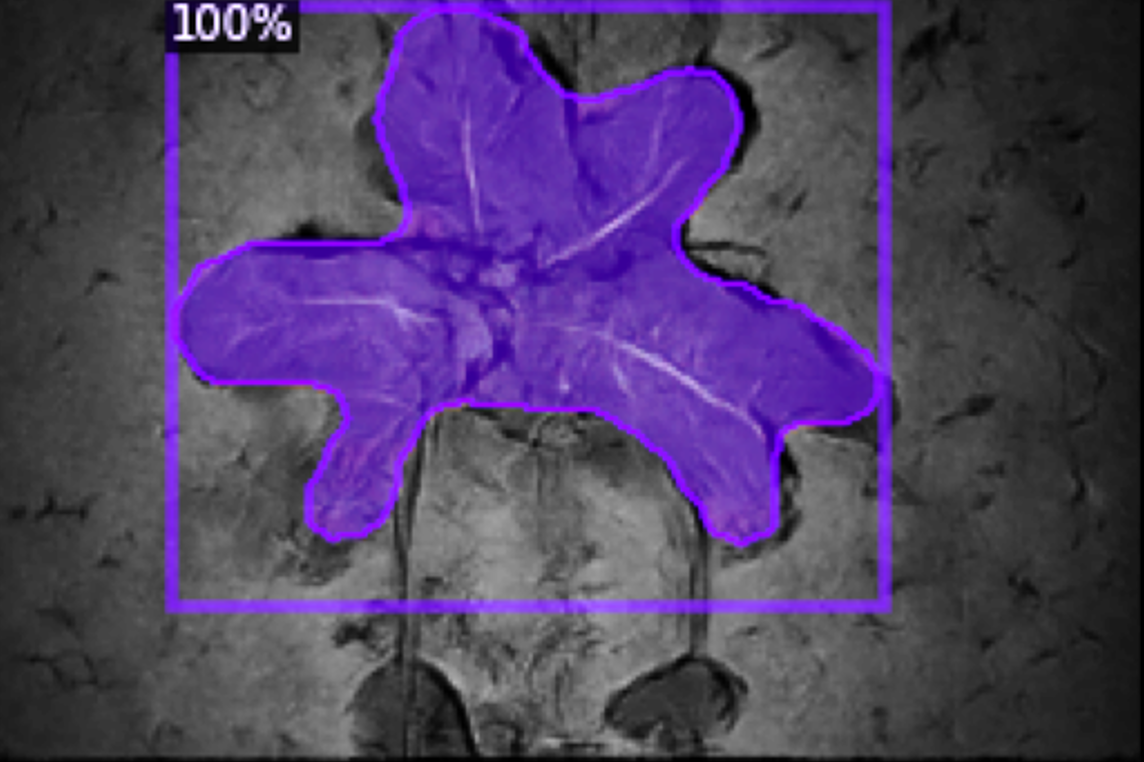}}
\subfloat[reference A]{\includegraphics[height=2.8cm]{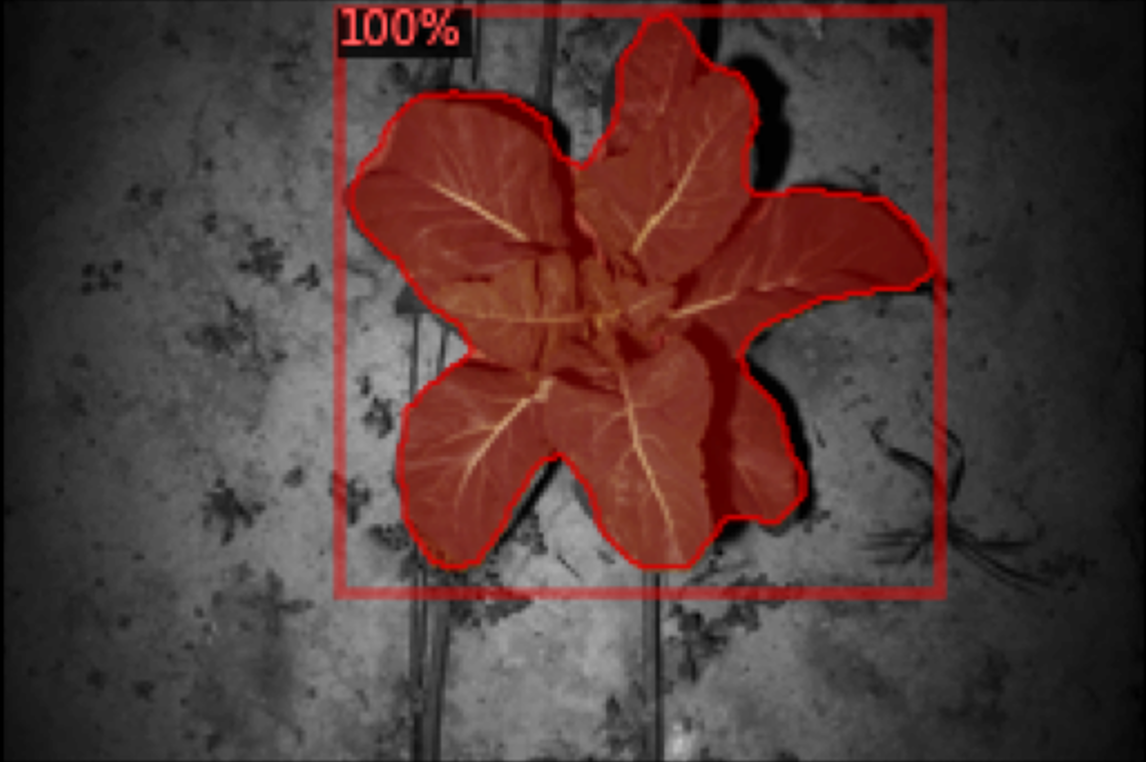}}

\subfloat[generated B]{\includegraphics[height=2.8cm]{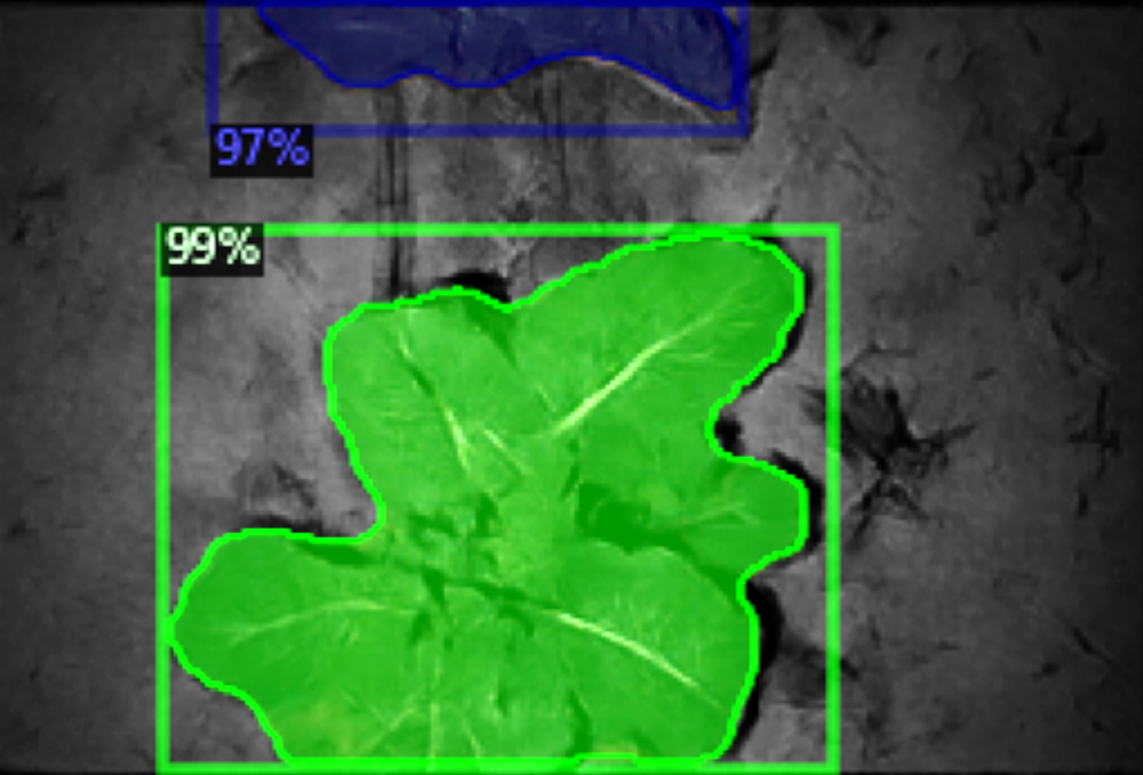}}
\subfloat[reference B]{\includegraphics[height=2.8cm]{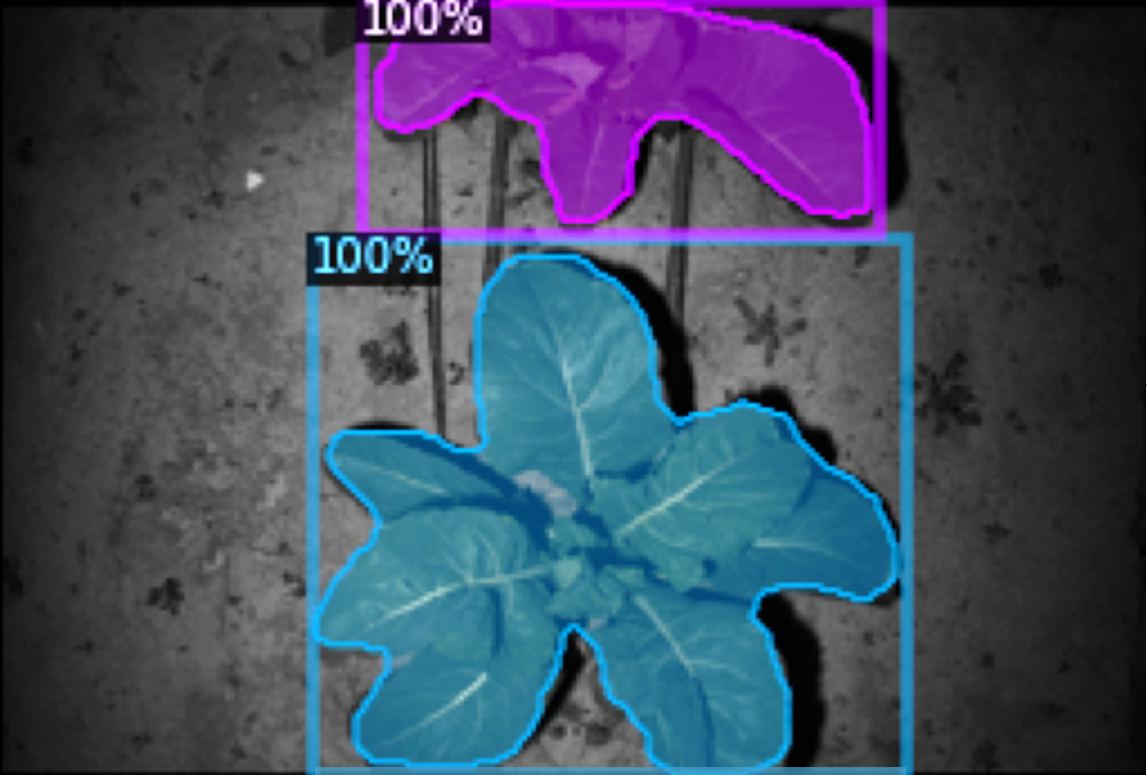}}
\caption{Two examples, A and B, of instance segmentation using Mask~R-CNN performed on two pairs of generated and associated reference images of cauliflower growth week 5. Colors are chosen randomly and have no meaning. The classification certainty is indicated in the corners of the boxes.}
\label{fig:cauli_instances}
\end{figure}

\subsection{Growth prediction pipeline}
The processing can be summarized in 3 steps: training, prediction and evaluation (see \figref{fig:pipeline}). First, a cGAN model is learned on aligned image pairs. In the second step, the generator is used to generate growth predictions from unseen images from early growth stages. Third, the generated images are evaluated, utilizing instance segmentation and \text{FID}. In the following, we give more details about the framework and the evaluation methods.

\subsubsection{Conditional GAN}
As conditional GAN we use the Pix2Pix framework provided by \cite{isola2017image}.
With this, a mapping $\ical{G}:\{\m{X}, z\} \rightarrow \m{Y}$ is learned, where \m{X} is the input image, $z$ is random noise, which is an indirect input realized as dropout within the architecture, and \m{Y} is the output image.
The cGAN consists of a generator \ical{G} and a discriminator \ical{D}.
It is called adversarial training because \ical{G} tries to generate images that are indistinguishable from real images, while \ical{D} is trained to identify the generated images of \ical{G} as such. 
In GAN literature, the generated images are also called fake or artificial images.
The objective function is therefore generally formulated as
\begin{equation}
\ical{G}^{*}=\arg \min _{\ical{G}} \max _{\ical{D}} \mathcal{L}_{\text{cGAN}}(\ical{G}, \ical{D}),
\end{equation}
with the loss function
\begin{equation}
\mathcal{L}_{\text{cGAN}}(\ical{G}, \ical{D})= \mathbb{E}_{X, Y}[\log \ical{D}(\m{X}, \m{Y})]+\mathbb{E}_{X, z}[\log (1-\ical{D}(\m{X}, \ical{G}(\m{X}, z)))],
\end{equation}
where \ical{G} tries to minimize it, while \ical{D} tries to maximize it.
If $\mathcal{L}_{\text{cGAN}}(\ical{G}, \ical{D})$ becomes minimal, either the generator is powerful, or the discriminator is very weak, and vice versa, if it becomes maximal.
Therefore, the training goal is to realize both adversarial goals at the same time so that they are balanced in the best case at the end of the training.
In order to achieve this, the gradient descent is always alternately applied first to the discriminator and then to the generator.
To additionally avoid blurring, the objective function is supplemented by L1-loss,
\begin{equation}\mathcal{L}_{\text{L1}}(\ical{G})=\mathbb{E}_{X, Y, z}\left[\|\m{Y}-\ical{G}(\m{X}, z)\|_{1}\right]\end{equation}
which forces greater similarities between the real and generated images.
This loss is added to the objective function, 
\begin{equation}
\ical{G}^{*}=\arg \min _{\ical{G}} \max _{\ical{D}} \mathcal{L}_{\text{cGAN}}(\ical{G}, \ical{D})+\lambda \mathcal{L}_{\text{L1}}(\ical{G})
\label{eq:pix2pixCmpltObjctv}
\end{equation}
where $\lambda$ is used to control the weighting of the losses.
Since noise in the input would be suppressed, the noise is implemented via dropout layers, and thus a stochastic result is obtained.

The generator \ical{G} network is a U-Net \cite{ronneberger2015u} with skip-connections.
Hereby, the input images are first processed in an encoder architecture until a bottleneck layer, to which a symmetrical decoder structure is attached.
Skip connections are used so that significant features of earlier layers, such as edges from which the position in the image or the plants' size can be determined, are not lost in the bottleneck.

The discriminator \ical{D} is a convolutional network that classifies the generated images into real and fake. 
A particular aspect is that the images are not processed in the network as a whole, but rather smaller patches of the input image, of which an average classification value is calculated.
This has the advantage of learning to model a more refined structure and a better texture instead of coarser structures \cite{isola2017image}.

Apart from parameter differences to \cite{isola2017image} described in more detail in \secref{sec:exp_setup} in subsubsection `cGAN Configuration', there are no modifications to the cGAN architecture of \ical{G} or \ical{D} to specialize it for our application.

\begin{figure}[t]
    \centering
    \includegraphics[width=1\columnwidth]{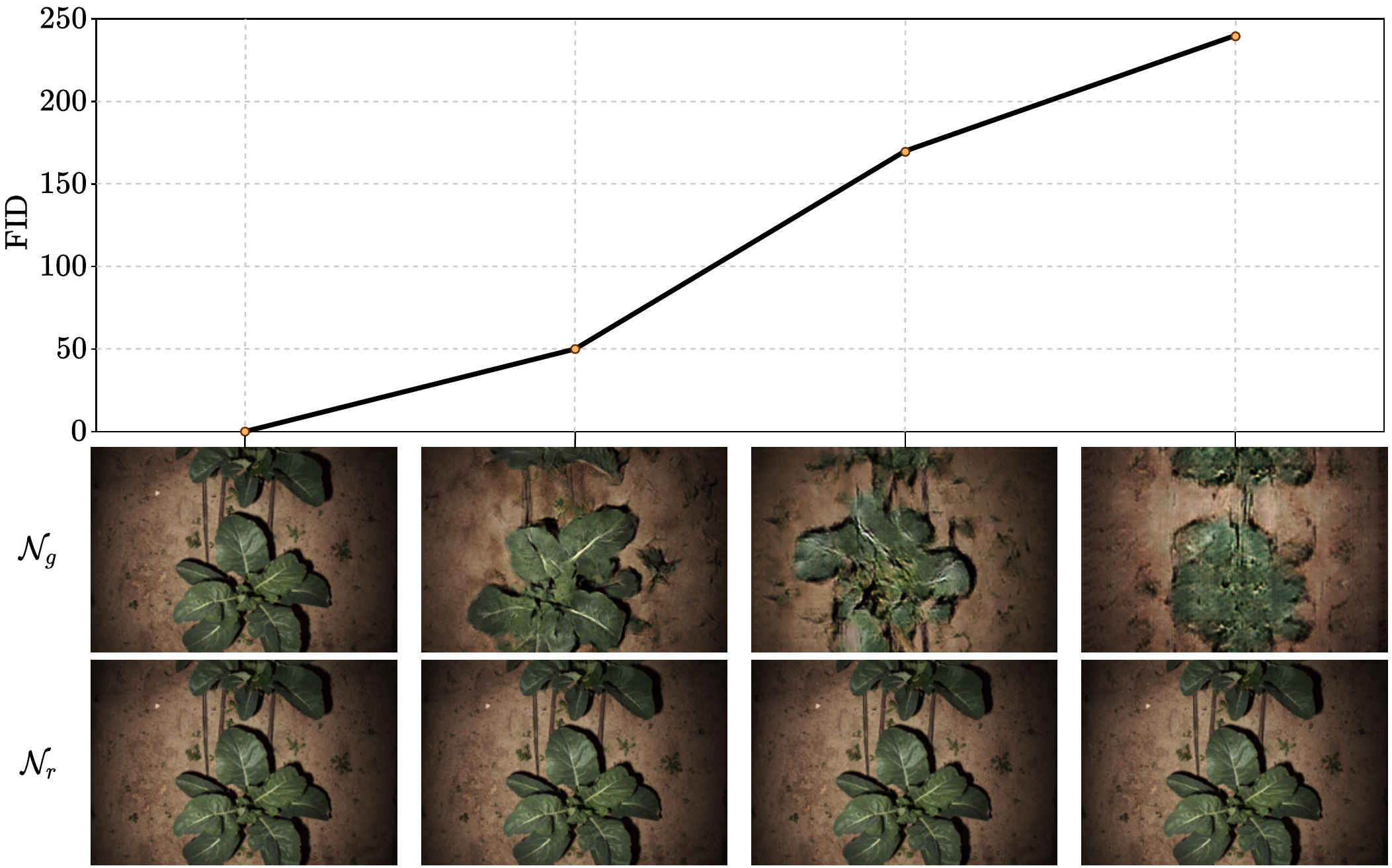}
    \caption{Illustration of \text{FID}($\mathcal{N}_r,\mathcal{N}_g$) in dependence of different input image distribution. Each image in the upper line represents a generated image distribution. Each image in the bottom line represents the associated reference image distribution. The higher the similarity between the distributions, the smaller the \text{FID}($\mathcal{N}_r,\mathcal{N}_g$).}
    \label{fig:fid}
\end{figure}

\subsection{Evaluation}
We perform the evaluation qualitatively and quantitatively, where we assess the appearance of single plants and the distribution of the generated plants. 

\subsubsection{Evaluation by instance segmentation}
The first part of the evaluation focuses on the appearance of the plants.
For this, manually derived segmentation masks of the real images are compared to estimated segmentation masks of the generated images by means of parameters such as extent and area.
Since semantic segmentation, where each pixel is assigned a class, is insufficient due to limitations when plants overlap, we utilize Mask~R-CNN to compute segmentation masks, which are semantic segmentations of each individual plant instance.
The instance segmentation masks can be used to quantify plant traits such as projected leaf area.
Besides the segmentation masks, we use estimated bounding boxes of the instances, from which we obtain the parameters diameter and center of the plant.

More specifically, for instance segmentation we use the Mask~R-CNN framework Detectron2 \cite{wu2019detectron2}.
It is pre-trained on everyday objects of the large scale COCO dataset \cite{lin2014microsoft} and fine-tuned on labeled images from the respective datasets.
The train and test data for fine-tuning include images from all growth stages.
For instance segmentation of \textit{A.~thaliana}, the Mask~R-CNN model is fine-tuned on about 850 images and evaluated on about 250 images.
For experiments with cauliflower of the Brassica dataset, it is fine-tuned with 25 images and evaluated on 10 images.
Although the amount of training data for fine-tuning in the Brassica dataset is small, it is sufficient because basic features are already learned in the comprehensive pre-training. 
A separate instance segmentation model is trained for each dataset, wherein both datasets we restrict ourselves to two classes, plant and background.
For both datasets, high-quality bounding boxes with an average precision $>~75~\%$ and semantic instance segmentation masks with an average precision $>~70~\%$ are estimated.
In all experiments, the same model is applied to the reference and the generated images (\figref{fig:cauli_instances}).

From the bounding boxes, the center position and the width of the plants are derived. 
Here, the center position is defined as the center of the leaf extent, which approximates the plant's actual center.
As width we refer to the extension of the cauliflower heads vertical to the plant row and thus vertical to the direction of motion of the robot.
In contrast to the height, which is defined here as the bounding box size in the direction of the plant row, the width is not affected by overlap errors, as the distance between rows is higher than between plants.
The segmentation area is used to determine the plants' size, i.e., the number of pixels covering the plant, which can be converted to a metric given a scale.

\subsubsection{Evaluation by Fréchet Inception distance}
The second part of the evaluation focus on the Fréchet Inception Distance (\text{FID}, \cite{heusel2017gans}), which calculates the distance between the Gaussian image distributions of the real images and the generated images.

The basic principle can be described in three steps:
First, a pre-trained Inception Net embeds both image sets to a new feature representation.
This feature representation is the activation of the flattened pooling layer with dimension 2048, one of the deepest layers in the Inception Net.
Although the Inception Net was trained with images from ImageNet instead of plant images, this layer represents the basic features of the image. 
Therefore it is suitable for embedding for a wide range of RGB datasets.
Second, Gaussian distributions $\mathcal{N}$ are defined by calculating the mean values $\d \mu$ and covariance matrices $\Sigma$ of these features.
In the last step, the Wasserstein-2 distance between the two Gaussian distributions is computed with
\begin{equation}
\text{FID}(\mathcal{N}_r,\mathcal{N}_g)=\left\|\d{\mu}_{r}-\d{\mu}_{g}\right\|_{2}^{2}+\operatorname{Tr}\left({\Sigma}_{r}+\Sigma_{g}-2\left(\Sigma_{r}
\Sigma_{g}\right)^{\frac{1}{2}}\right),
\end{equation}
where $\mathcal{N}_r(\mu_{r},\Sigma_{r})$ and $\mathcal{N}_g(\mu_{g},\Sigma_{g})$ are the Gaussian distributions of reference and generated images.
\figref{fig:fid} illustrates how \text{FID}($\mathcal{N}_r,\mathcal{N}_g$) value behaves in different constellations of input distributions.
The smaller the \text{FID}($\mathcal{N}_r,\mathcal{N}_g$), the higher the similarity between reference and generated distributions.

\section{Experiments and Results}\label{sec:resuls}
We performed two different experiments denoted by \texttt{Arabidopsis} and \texttt{Cauliflower} by calculating a different cGAN-based growth model for each data set.
We first describe the different experimental setups and then analyze and compare the respective results.

\subsection{Experimental Setup}\label{sec:exp_setup}

\subsubsection{Experimental goals}

In the experiment \texttt{Arabidopsis} we use the Aberystwyth leaf evaluation dataset and calculate a model that predicts the phenotype of \textit{A.~thaliana} 17 days into the future.
The aim is to train a model, which predicts early growth stages (e.g., day 21 $\rightarrow$ 38) and late growth stages (e.g., day 38 $\rightarrow$ 55) simultaneously.

Similarly, in \texttt{Cauliflower} we use the Brassica dataset and calculate a model that predicts the phenotype of cauliflower three weeks into the future.
The model allows growth predictions of plants with any growth stage between weeks 1 and 6 after planting as the condition.
Accordingly, the results should be realistic plants with a growth stage between weeks 4 and 9.

We have different expectations of the generated images in terms of realistic appearance, reliability, and reasonableness:
\begin{itemize}
    \item Realistic appearance: 
    The images as a whole should show details, be as blur-free as possible, and contain no artifacts. Plants should look natural in terms of color, structure, and size.
    \item Reliable generation: The model should have a generalization ability, i.e., the generation should not only work for a part of the dataset but be robust to different growth stages, shapes of the plant, and background conditions.
    \item Reasonable output: The generated image should not be arbitrary but depend on the input image. If a model is trained on different field treatments, it should be able to predict different plant sizes accordingly.
\end{itemize}

It must be noted that we do not expect the generated image to contain every detail of the reference image.
Especially the orientation and size of single leaves varies strongly between growth stages that are 17 days or three weeks apart and therefore cannot be reconstructed.
Rather, we expect to produce a plant that is similar to the reference in terms of overall size and position in the image, and thus geo-referencing in the field to be within the accuracy of the image alignment.
It is also expected that the highly accurate image alignment and almost continuous observation positively impact the results of \texttt{Arabidopsis} compared to \texttt{Cauliflower}.

\subsubsection{Data preparation}
The \texttt{Arabidopsis} train-test split is realized on the basis of a clear spatial separation. 
Three sets, namely tray031, tray032, and tray033, are used for training and tray034 for testing.
After cutting individual plants from the trays over all timestamps and building the aligned image pairs, it results in 7618 training pairs and 2707 test pairs.

For the \texttt{Cauliflower} dataset, we perform a spatially disjoint train-test split such that for training the plants from bed 01 are used, for testing the plants from bed 03.
A distinction between irrigated and fertilized treatments is not made during training, so we use the cauliflower plants from the entire bed 01.  
For testing the 3-week aligned image pairs of bed 03 are distinguished between the field treatments \{\texttt{i+f+}, \texttt{i+f-}, \texttt{i-f+}, \texttt{i-f-}\}.
In this way, we intend to investigate whether the model covers different growth behavior with modified fertilization and irrigation.
The number of aligned image pairs for this experiment is recorded in \tabref{tab:w3nums}.

\input{tables/w3_aligned_num.tex}
\subsubsection{Data pre-processing}
For training, we clean the aligned image pairs in two aspects.
On the one hand, we sort out pairs of images where no plant is visible in domain A but is visible in domain B.
This occurs mostly in the early growth stages of experiment \texttt{Cauliflower} when an image is taken between two plants, but the plant in domain B is already so large that it grows into the image.
On the other hand, we sort out images where a plant is visible in domain A but not in domain B.
This case occurs when the plant is harvested in the meantime. 
Since the images were taken on a trial field, this happens often, but such image pairs would falsify the training.
However, image pairs in which the plant is only partially visible in both domains are not rejected. 
Instead, they contribute to making the training more robust by increasing diversity in the data distribution.

\begin{figure*}[t]
\centering
\captionsetup[subfigure]{labelformat=empty}
\subfloat[]{\includegraphics[height=4.0cm]{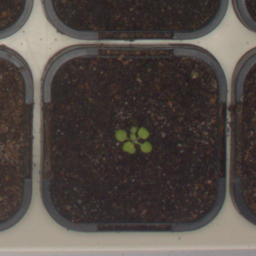}
}
\subfloat[]{\includegraphics[height=4.0cm]{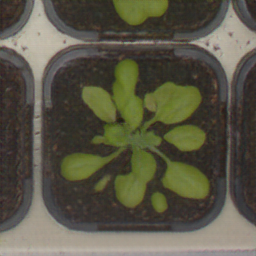}
}
\subfloat[]{\includegraphics[height=4.0cm]{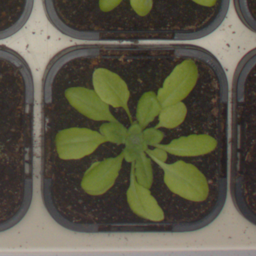}
}

\subfloat[]{\includegraphics[height=4.0cm]{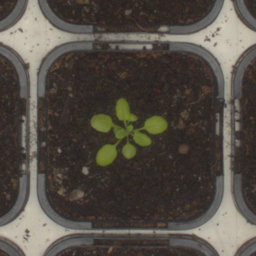}
}
\subfloat[]{\includegraphics[height=4.0cm]{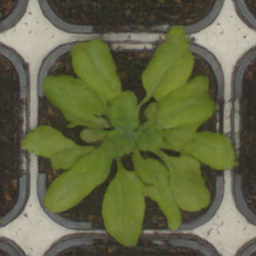}
}
\subfloat[]{\includegraphics[height=4.0cm]{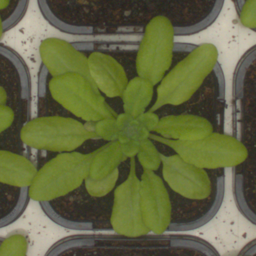}
}

\subfloat[domain A: test-input]{\includegraphics[height=4.0cm]{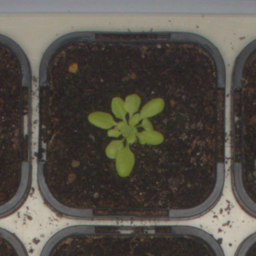}
}
\subfloat[domain B: generated]{\includegraphics[height=4.0cm]{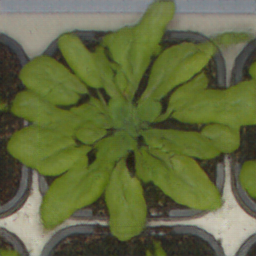}
}
\subfloat[domain B: test-reference]{\includegraphics[height=4.0cm]{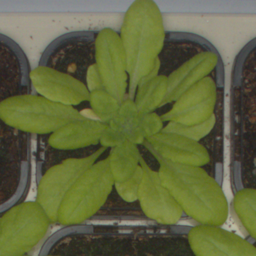}
}
\caption{Some example results of experiment \texttt{Arabidopsis}. The left column represents the test-input in domain A, the middle column is the generated output in domain B, and the right column shows the aligned test-reference image to domain A in domain B. A prediction of early growth stages (top row: day after sowing 23 $\rightarrow$ 40), middle stages (middle row: day after sowing 30 $\rightarrow$ 47) and advanced growth stages (bottom row: day after sowing 37 $\rightarrow$ 54) is possible.}
\label{fig:d17_example}
\end{figure*}

\subsubsection{cGAN configurations}
Since the generator is optimal for square images, all rectangular image pairs are provided with an equally large black border on the top and bottom sides.
Data augmentation consists of random cropping, vertical and horizontal mirroring, and rotations applied to both domains of the train and test image pairs. 
The rotation options are limited to $0\deg$ and $180 \deg$ to maintain the geometry of the vertical alignment of the cauliflower rows in each image.
All images are reduced to the same size of $\SI{256}{\px} \times\SI{256}{\px}$ for training and testing.
The network architecture is maintained in its original state as presented in \cite{isola2017image}; however, some hyperparameters are adjusted.
For instance, the learning rate is $1e-4$, the loss weighting parameter $\lambda$ is 100, and the batch size is 1. 
The number of epochs is 160 for \texttt{Cauliflower} and 40 for \texttt{Arabidopsis}, in each case the second half with linearly decaying learning rate.

\subsection{Results and Discussions}

\subsubsection{Visual analysis of generated A.~thaliana images}
\figref{fig:d17_example} shows 3 examples of visual results of temporal prediction in the \texttt{Arabidopsis} experiment.
The upper row shows the prediction from day 23 to 40, the middle row from day 30 to 47, and the lower row from day 37 to 54.
It is clear to see that the prediction is successful in both early and late epochs because the generated images are highly similar to the reference images, both in terms of the extent and the number of leaves.
There are only little details that reveal the artificiality of the generated plants.
For instance, in the generated image on day 40 (upper row, center), there are two leaves in the lower part that are not attached to the plant with a petiole, and on day 54 (lower row, center), there is a small artifact in the upper right corner.
Likewise, when analyzing the leaf structures, one notices that in all images less petioles have unusual curvatures and some leaves have uncommon shapes (middle row, center).
In terms of color, the generated images are very natural, as the leaves take on slightly different shades of green, which can be all found in the reference and become slightly lighter from the inside out. 
Even yellowish leaves or parts of leaves can be found sporadically in the reference as well as in the generated image.
In general, it is noticeable that larger outer and inner smaller leaves are generated without being blurry.
Another noteworthy aspect is the detailed generation of the background. 
It looks nearly the same as the input, even the small lumps of dirt change their position in the generated image just like in reality.

\subsubsection{Evaluation of generated A.~thaliana images using Mask~R-CNN}
\begin{figure}[t]
    \centering
    \includegraphics[width=0.9\columnwidth]{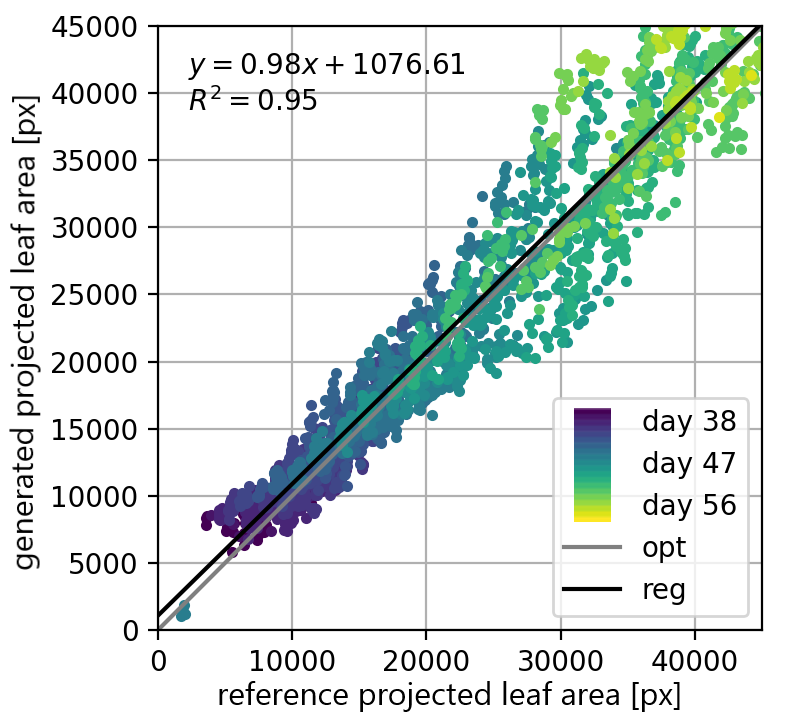}
    \caption{Comparison of the projected leaf area [in pixels] of reference and generated \textit{A.~thaliana} plants. In the scatter plot, one dot refers to a pair of reference and generated plant. The grey line indicates the optimal line, while the black line represents the regression line. In the upper left corner, the straight-line equation and the $R^2$ value are indicated.}
    \label{fig:d17_scatter}
\end{figure}

\begin{figure}[t]
    \centering
    \includegraphics[width=1\columnwidth]{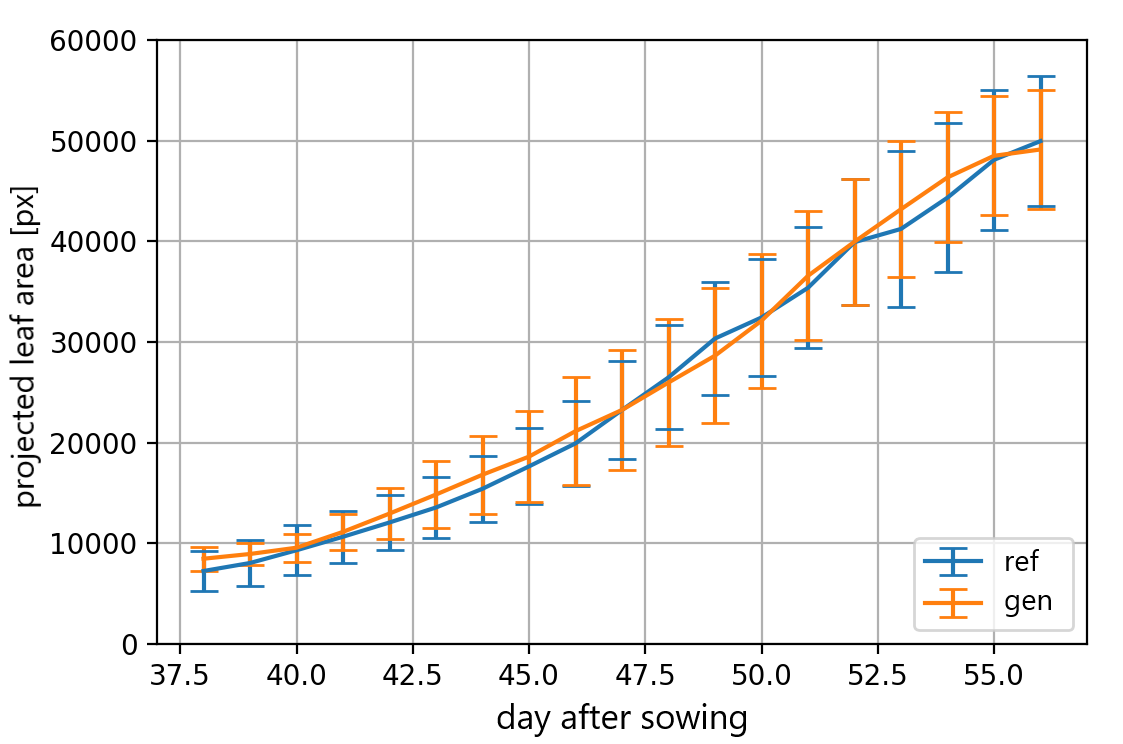}
    \caption{Measured and estimated growth of \textit{A.~thaliana} plants over time in pixels quantified as projected leaf area of reference and generated images. Only the prediction in the range from 38 to 56 days after sowing is displayed. The error bars indicate the standard deviation.}
    \label{fig:d17_weeks}
\end{figure}

\figref{fig:d17_scatter} and \figref{fig:d17_weeks} show the comparison of the generated and the reference images by means of the projected leaf area, where \figref{fig:d17_scatter} focuses on single plants and \figref{fig:d17_weeks} addresses the average values per week. 

\figref{fig:d17_scatter} shows a high correlation between the size of the projected leaf area of corresponding generated and reference images of the same domain. 
The high $R^2$ value of $0.95$ indicates the good performance of the model for temporal prediction of plant sizes.
The projected leaf area is derived from the semantic \textit{A.~thaliana} masks of the instance segmentation, which was obtained by Mask R-CNN for both the generated and reference images.
The color indicates sequentially the time of prediction, from dark blue (early growth phase) to green (medium growth phase) to yellow (late growth phase).
Only a slight overestimation of the projected leaf area can be seen over the whole period, which becomes smaller with increasing plant size (gradient $0.98$).

The quality of the temporal prediction of \textit{A.~thaliana} is underlined by \figref{fig:d17_weeks} when comparing the total size of reference and generated plants in the reference period (days 38 to 56).
The average size of the projected leaf area in the generated and reference images is almost the same for every day of the temporal prediction.
However, the generated curve is very smooth, while the reference curve has small bumps that are very typical of a true plant growth curve.
The maximum deviation is $\SI{2000}{\px}$ on day 54 and the average deviation over all days is less than $\SI{500}{\px}$, so remarkably low in all growth stages.
At later growth stages, the standard deviation of both, reference and generated plant, increases absolutely, showing increased plant size variability in advanced growth stages.

\subsubsection{Visual analysis of generated cauliflower images}
\begin{figure*}[p]
\centering
\captionsetup[subfigure]{labelformat=empty}
\subfloat[]{\includegraphics[height=3.7cm]{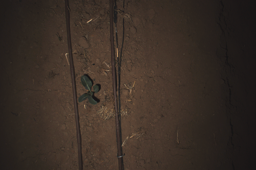}
}
\subfloat[]{\includegraphics[height=3.7cm]{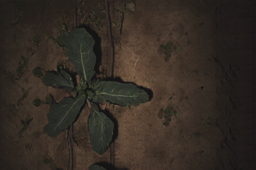}
}
\subfloat[]{\includegraphics[height=3.7cm]{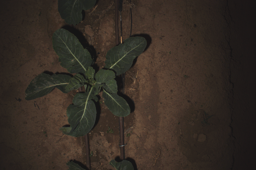}
}
\vspace{-0.6cm}
\subfloat[]{\includegraphics[height=3.7cm]{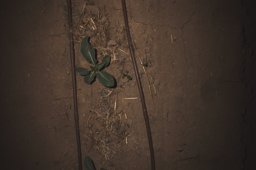}
}
\subfloat[]{\includegraphics[height=3.7cm]{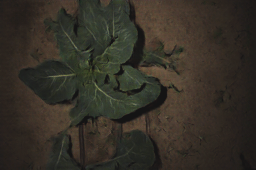}
}
\subfloat[]{\includegraphics[height=3.7cm]{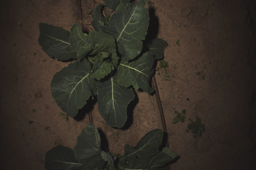}
}
\vspace{-0.6cm}
\subfloat[]{\includegraphics[height=3.7cm]{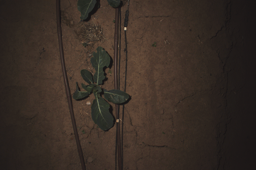}
}
\subfloat[]{\includegraphics[height=3.7cm]{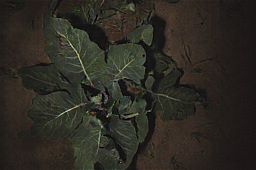}
}
\subfloat[]{\includegraphics[height=3.7cm]{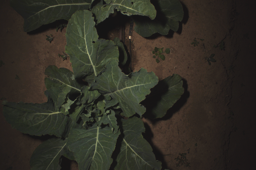}
}
\vspace{-0.6cm}
\subfloat[]{\includegraphics[height=3.7cm]{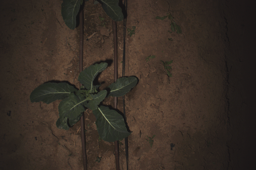}
}
\subfloat[]{\includegraphics[height=3.7cm]{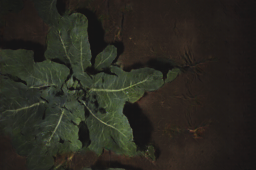}
}
\subfloat[]{\includegraphics[height=3.7cm]{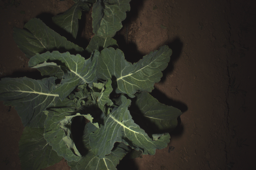}
}
\vspace{-0.6cm}
\subfloat[]{\includegraphics[height=3.7cm]{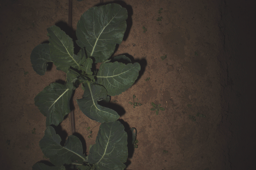}
}
\subfloat[]{\includegraphics[height=3.7cm]{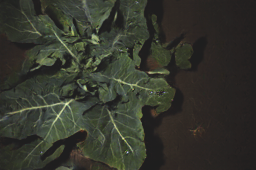}
}
\subfloat[]{\includegraphics[height=3.7cm]{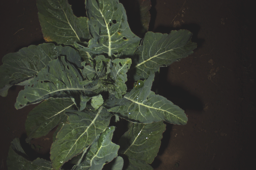}
}
\vspace{-0.6cm}
\subfloat[domain A: test-input]{\includegraphics[height=3.7cm]{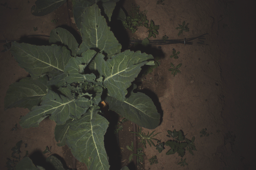}
}
\subfloat[domain B: generated]{\includegraphics[height=3.7cm]{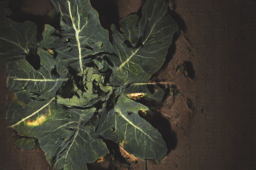}
}
\subfloat[domain B: reference]{\includegraphics[height=3.7cm]{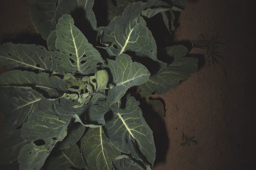}
}

\caption{Examples of input, generated and reference images of \texttt{Cauliflower} from week 1-6 for domain A and 4-9 for domain B. The rows show different growth prediction steps. Week 1 $\rightarrow$ 4 in the top row, week 2 $\rightarrow$ 5 in the second row up to week 6 $\rightarrow$ 9 in the bottom row. The left column represents the input domain A, the middle column is the generated output in domain B, and the right column shows the aligned reference image to domain A in domain B. The more similar the middle and right columns, the better the generated image.}
\label{fig:w3_example}
\end{figure*}

In the experiment \texttt{Cauliflower}, first, the appearance of the test images of different growth stages are visually assessed \figref{fig:w3_example}).
The generated cauliflower images of domain B (middle column) look realistic and could be mistaken for real cauliflowers by an unbiased judgment. 
The comparison of the generated images with reference images for every week (right column) shows a variety of reasons for this.
Although there is some noise in some locations, like on the left side of the cauliflower in the generated image of week 9, the overall sharpness of the generated images is almost as good as with the reference.
In addition, brightness, contrast, and saturation, as well as color values of foreground and background, match the reference.

A detailed look at the foreground shows that the size, number, and shape of the leaves is plausible. 
Apart from a few exceptions (bottom leaf in week 5, rightmost leaf in week 8), the orientation of the leaves towards the center of the plant is correct, which can be seen from the direction of the leaf veins.
Likewise, the image background is realistically represented.
Exceptional brightness levels like in week 8, in which the background is much darker than in the other images, are captured as well as small details in the background. 
For example, in the steps 1 $\rightarrow$ 4 and 2 $\rightarrow$ 5, the drainage pipes are visible in the generated image at reasonable positions.
In the same way weeds of various sizes are visible in the background next to the cauliflowers in all stages of growth.

When analyzing the relations between the generated images (middle column) and the reference image of input domain A (left column), which is used as a condition for the prediction, we observe a clear correlation for overall cauliflower size. 
A specific input size in domain A causes a certain output size in domain B, which matches the non-linear growth of cauliflower in the left column.
From week 1 to 3, cauliflower shows a rather slow growth, which picks up substantially by week 5, when plants show a fast increase of leaf growth and number. 
This rather exponential growth provides a challenge for prediction, as the relation between condition and expected output is not constant over time.
However, the resulting sizes in generated images of domain B (middle column) still fit the reference (right column) well.

Looking at the orientation of the individual leaves, there is not obvious pattern between reference images of domains A and B. 
Although the growth direction and orientation of individual leaves does not change within three weeks, emerging leaves often become more dominant and overlay other plant organs.
Therefore, our temporal growth prediction considers the plant's development as a whole, rather than changes in individual leaves.
To make use of the generated images, it is essential that the center of plants does not change and is well geo-referenced, just as is expected for plants as sessile organisms. 
Our example images show that despite inaccuracies of the image alignment (see \secref{sec:data}), the center position of the plants in the left and middle column match very well between both domains.

\subsubsection{Evaluation of generated cauliflower images using Mask~R-CNN}
\begin{figure*}[tbh]
\captionsetup[subfigure]{labelformat=empty}
\centering
\subfloat[treatment \texttt{i-f-}]{\includegraphics[height=7.0cm]{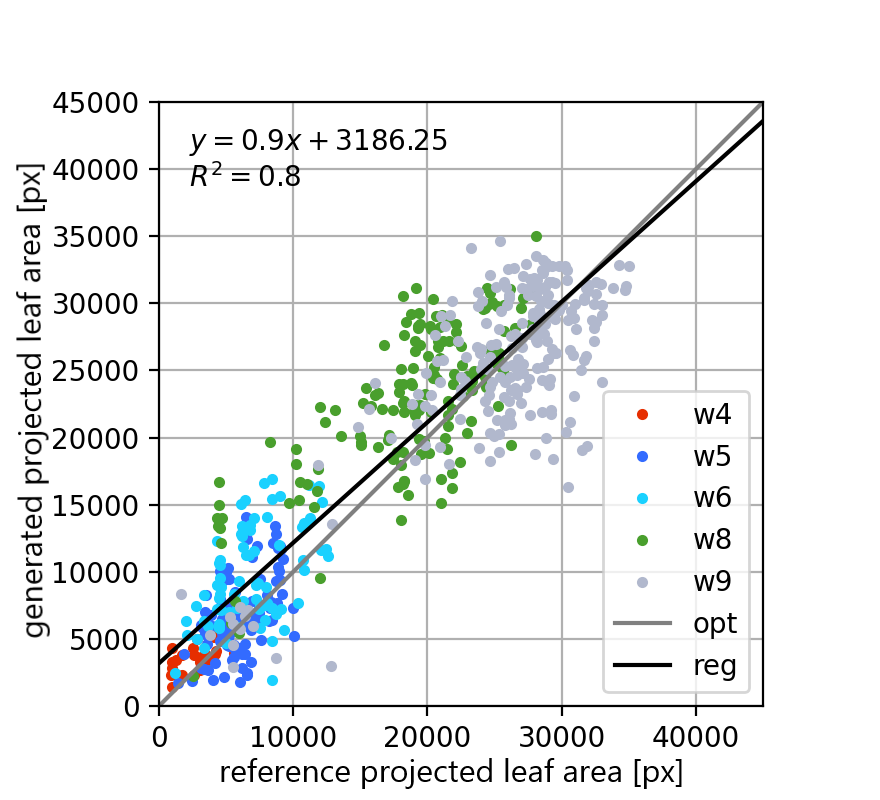}}
\hspace{-2mm}
\subfloat[treatment \texttt{i-f+}]{\includegraphics[height=7.0cm]{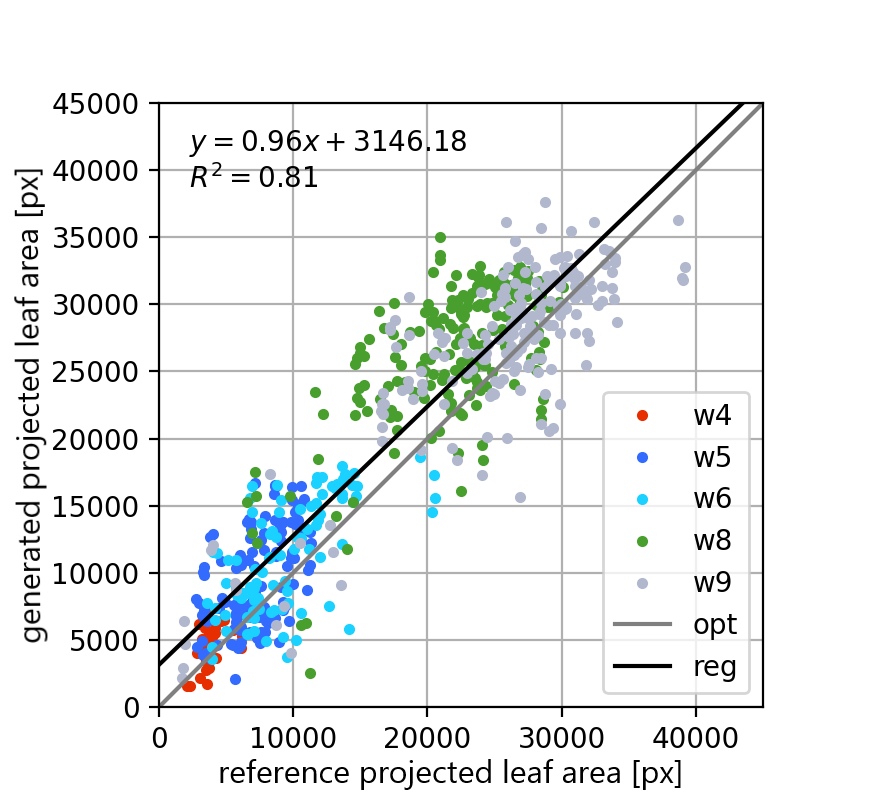}}
\vspace{-2mm}

\subfloat[treatment \texttt{i+f-}]{\includegraphics[height=7.0cm]{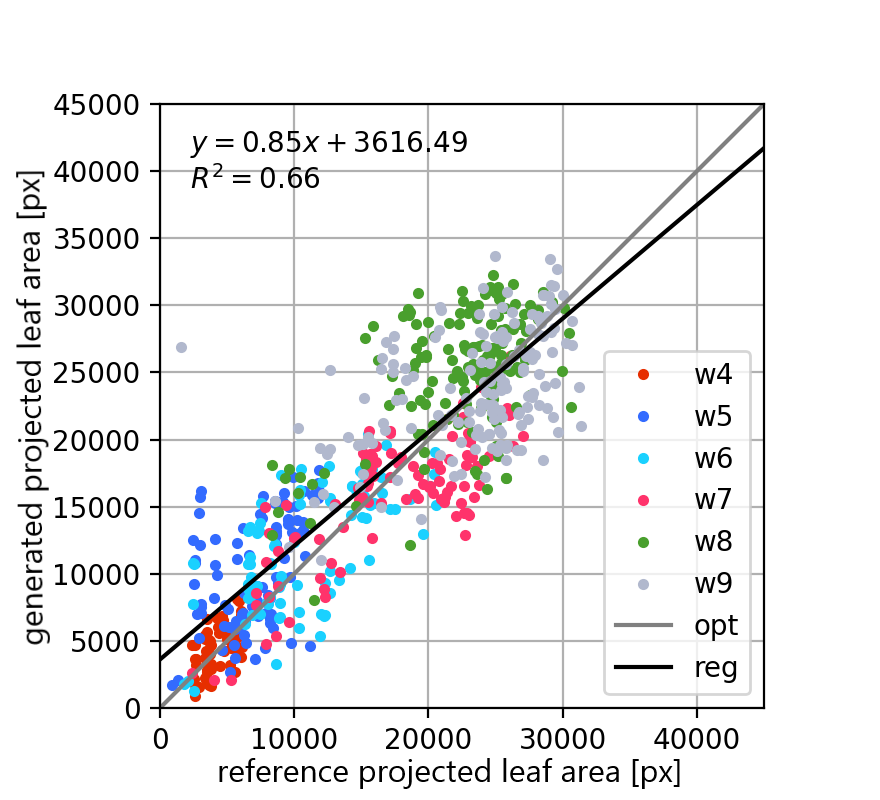}}
\hspace{-2mm}
\subfloat[treatment \texttt{i+f+}]{\includegraphics[height=7.0cm]{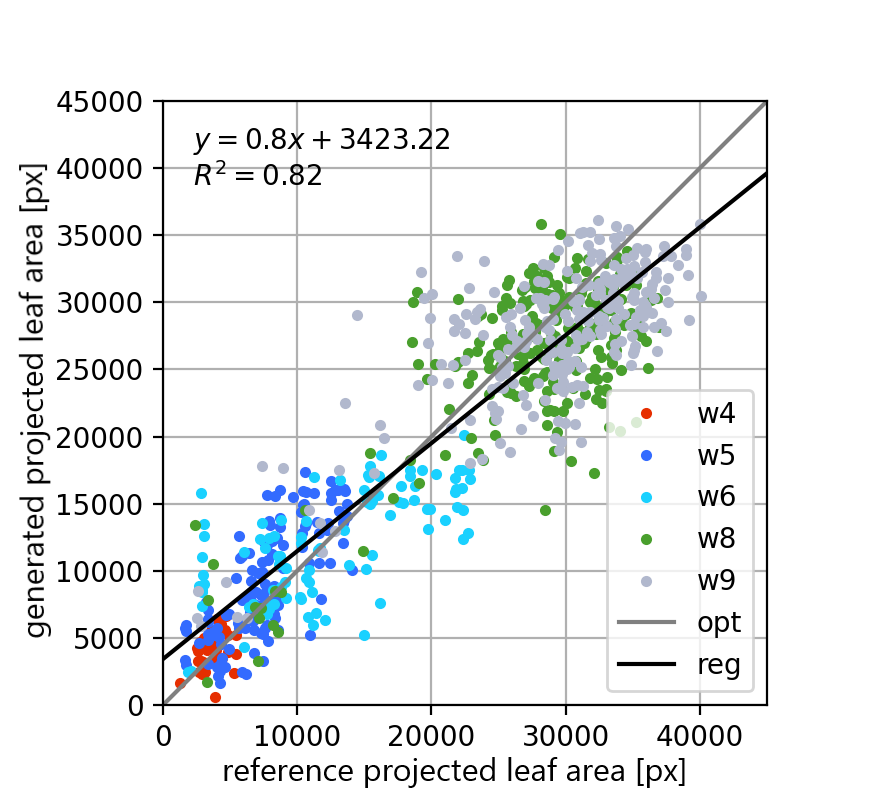}}
\caption{Comparison of the projected leaf area in  pixels of reference and generated cauliflower images. The data are separated into four subplots according to their irrigation (\texttt{i}) and fertilization (\texttt{f}). One dot in the scatter plots refers to a pair of reference and generated plant, where the color of the dots indicate the week (w). The grey line indicates the optimal line, while the black line represents the regression line. In the upper left corner, the straight-line equation and the $R^2$ value are indicated. }
\label{fig:w3_scatter}
\end{figure*}

In \figref{fig:w3_scatter}, we compare the growth of generated (y-axis) and reference instances (x-axis) for different treatments.
For all four treatments, it is visible that the points cluster in an area around the grey lines, which is the target relationship, where the generated projected leaf area is identical to the reference one.
Overall, there is a trend that plants in early growth stages are predicted to grow a little too large (black regression line above the optimal grey line), and plants later in growth tend to grow a little too small (black below the grey line).
This is also evident from the derived equation of the fitted line, which has a gradient smaller than 1 in all cases.
This is likely caused by the observed exponential growth, while the model is trained to work with all plant ages.
However, $R^2$ values from 0.66 to 0.82 show that temporal prediction works well despite the field conditions.
It is also noticeable that points of some weeks are not clearly separated.
For instance, week 5 + 6 and week 8 + 9 partially overlap.
This is due to the natural variance in the expression of the plants' phenotype - even cauliflowers exposed to the same field treatments develop differently within a certain range.
It is also noticeable that the dispersion of the points increases with the plant age, which is explained by the higher natural variance with rising projected leaf area.

It is apparent that well irrigated treatments \texttt{i+f-} and \texttt{i+f+} are more underestimated than less irrigated treatments \texttt{i-f+} and \texttt{i-f-}, which can be seen from the amount of points below the grey line.
This comes from the joint training with all field treatments, which influence that different treatments counteract each other, so that there is a shift in the learned generator towards the average growth of all treatments.
Therefore large plants are estimated slightly too small, and small plants slightly too large.

\begin{figure*}[thb]
\captionsetup[subfigure]{labelformat=empty}
    \centering
    \subfloat[Weekly development of reference plants]{        \includegraphics[width=0.48\textwidth]{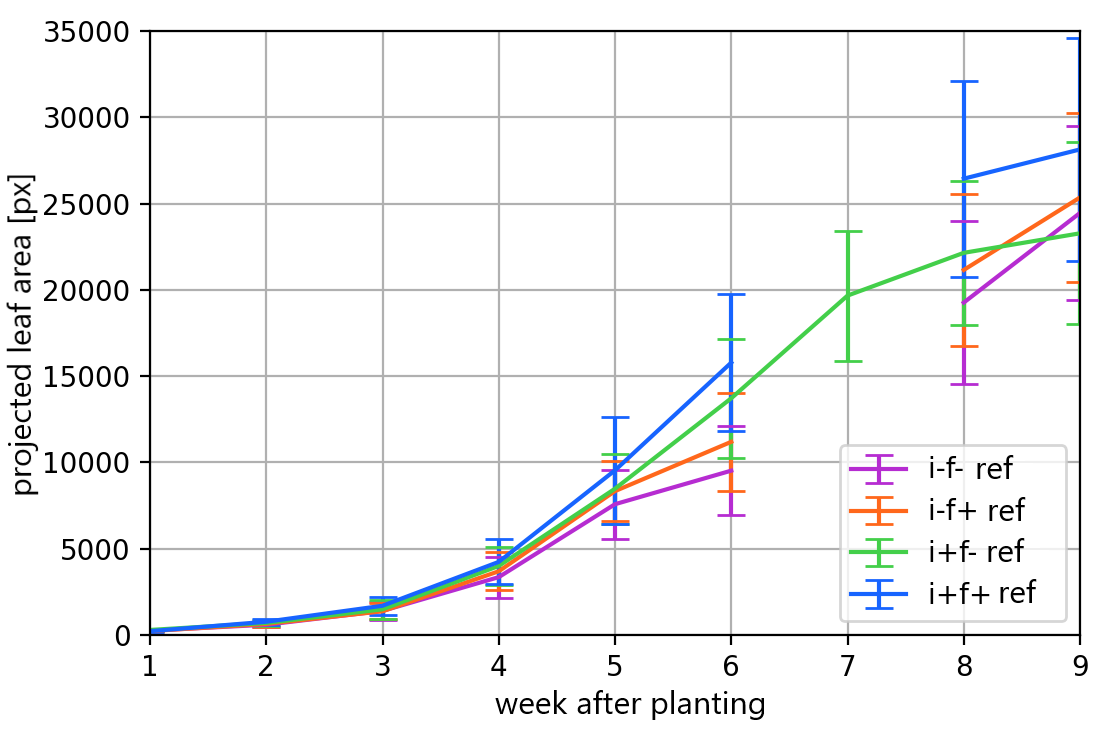}}
    \subfloat[Weekly development of generated plants]{        \includegraphics[width=0.48\textwidth]{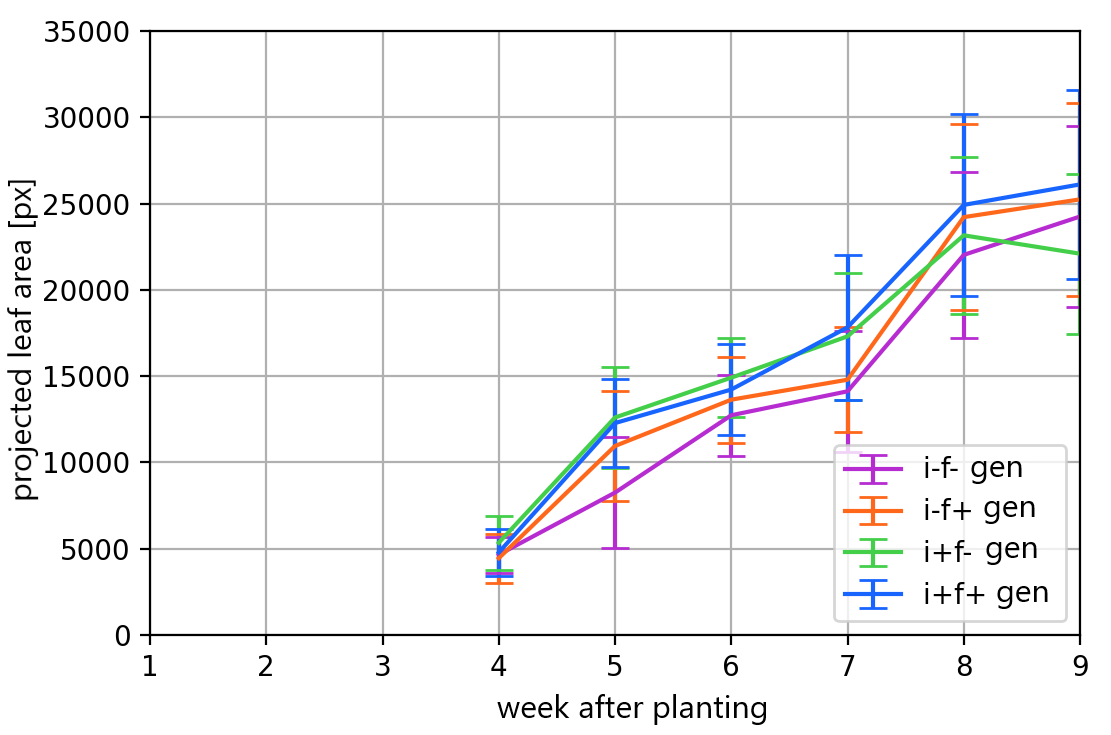}}
    \caption{Size comparison of the mean segmentation area [in pixels] over weeks between reference (left) and generated images (right) of experiment \texttt{Cauliflower}. The error bars indicate the standard deviation in the respective weeks. The lines are separated into the different field treatments.}
    \label{fig:w3_weekstats}
\end{figure*}

Nevertheless, the absolute differences in size between the treatments is well modeled, which is best seen in \figref{fig:w3_weekstats}.
We analyzed the mean projected leaf area, which is equal to the mean segmentation area, along with the standard deviation of the reference plants (left) and the generated plants (right) weekly.
For these experiments, only those plants are taken into account that are located in the image center, as indicated by the center of the bounding box.
The reason is to avoid plants that are not completely visible in the image and influence the distribution of the sizes.

It is clearly seen from week 5 onward that the sizes of the reference plants are strongly dependent on the field treatments.
Well irrigated and fertilized plants grow better than plants lacking water and nutrients, which is in line with expectations and analyses performed by \cite{bender2020high}.
The size of reference plants grown under \texttt{i+f+} (blue line) was bigger compared to \texttt{i+f-} (green), followed by \texttt{i-f+} (orange) and \texttt{i-f-} (magenta).
Although the values of the generated plants with \texttt{i+f-} treatment is larger than \texttt{i+f+} in weeks 4 to 6, there is a clear analogy to the growth of reference plants under the respective treatments.
Noteworthy in week 9 is the bending of the green line in both reference and generated plants.
Whatever hampered plant growth at this later stage, it was apparently already encoded in the images from week 6 and detected by the model, although plant size was not different affected at this time.

In week 7, the generated plants show a smaller increase in size in comparison to the other weeks.
Using the growth pattern of the reference plants with \texttt{i+f-} treatment for comparison, one would expect the generated plants in week 7 to be about $\SI{2000}{\px}$ to $\SI{3000}{\px}$ larger in all field treatments.
We see two reasons that it does not occur and that plants are often underestimated in week 7. 
First, the training data for step 4 $\rightarrow$ 7 (see \tabref{tab:w3nums}) is missing and the generator incorrectly interprets some input images from week 4 as images from week 3.
Second, the beginning of exponential growth at these growth stages causes difficulties, as small differences in the condition have large effects on the generated images.
We assume that more training data from the exponential growth period, at best under different climatic conditions, would improve this behaviour.
However, in all weeks, the sizes of the generated cauliflowers are within the standard deviation of the reference cauliflower sizes.

\subsubsection{Analyzing \text{FID} scores}
Besides evaluating the quality of single plant instances, we further calculate the \text{FID} scores for both experiments to assess the accuracy of the estimations at each growth stage.
Unlike many related works, we analyze three different FID scores, namely
\begin{itemize}
    \item \text{FID}($\mathcal{N}_r,\mathcal{N}_g$): similarity of the distributions of test-reference and generated images, 
    \item \text{FID}($\mathcal{N}_r,\mathcal{N}_t$): similarity of the distributions of test-reference and training-reference images,
    \item \text{FID}($\mathcal{N}_g,\mathcal{N}_t$): similarity of the distributions of generated and training-reference images.
\end{itemize}
The results are summarized in \tabref{tab:fid}.

\input{tables/fid_table.tex}

An absolute evaluation of these values is not trivial, but two aspects are noteworthy for \texttt{Cauliflower} experiment. 
First, as expected, we observe that the generated images have a higher average similarity to the test-reference images than to the train-reference images due to different growth-stages.
Second, we observe that the distributions of generated and test-reference images are more similar to each other than the distributions of train-reference and test-reference because \text{FID}($\mathcal{N}_r,\mathcal{N}_g$) is lower than \text{FID}($\mathcal{N}_r,\mathcal{N}_t$).
We see this as evidence that \text{FID}($\mathcal{N}_r,\mathcal{N}_g$) compares distributions that closely resemble and that our neural network-based growth model produces reasonable images with a realistic appearance. 

In contrast to the \texttt{Cauliflower} experiment, in the \texttt{Arabidopsis} experiment, \text{FID}($\mathcal{N}_r,\mathcal{N}_g$) is larger than \text{FID}($\mathcal{N}_r,\mathcal{N}_t$). 
While \text{FID}($\mathcal{N}_r,\mathcal{N}_g$) only slightly exceeds the mean value in this category of the experiment \texttt{Cauliflower}, \text{FID}($\mathcal{N}_r,\mathcal{N}_t$) has decreased significantly in direct comparison. 
Thus, the quality of the generated images is not worse than in the \texttt{Cauliflower} experiment, but the distributions of test-reference and training-reference have an exceptionally high similarity due to the laboratory conditions.
Again, \text{FID}($\mathcal{N}_r,\mathcal{N}_g$) is smaller than \text{FID}($\mathcal{N}_g,\mathcal{N}_t$).
This is essential as it suggests that in both experiments the plants are actually generated from the input conditions rather than replicating the best fitting training pattern.

\subsubsection{Processing times}
Following the pipeline in \figref{fig:pipeline}, the runtime for each step has to be considered separately. 
To train a suitable generator, the computing time is about $\SI{5}{\min}$ per epoch for \texttt{Arabidopsis} and about $\SI{3.5}{\min}$ per epoch for \texttt{Cauliflower}, whereas this time is largely determined by the dataset size and the number of augmentations.
The convergence time of GAN training depends mainly on how diverse the distribution of images is, which is higher for \texttt{Cauliflower} than for \texttt{Arabidopsis}, requiring more epochs overall (160 instead of 40).
The two following inference steps, namely the application of the generator to predict future images and the calculation of the projected leaf area over pre-trained Mask R-CNN are real-time capable.

\subsubsection{Key influencing factors for generated images in the field}
We identified four factors from both experiments on which the accuracy of the temporal predictions mainly depend, which is essential for applications in the field: 1) an observation rate as frequent as possible and thus a large number of aligned images with different time references to train the model; 2) an exact geo-referencing of the images, which in this case was produced by laboratory conditions, but can also be achieved in the field; 3) high-quality images with sufficient spatial resolution to detect nutrient, water or other deficiencies at an early stage; and 4) a complete view of whole plants with little overlap of neighboring plants.

\subsubsection{Scalability and Limitations}
There are a number of different types of empirical and process-based crop growth models that incorporate a variety of environmental and management factors \cite{di2016overview}. 
Typically, those knowledge-based models include influencing variables explicitly via input parameters into the model, rather than implicitly via images.
While knowledge-based models allow flexible simulations if they are based on physical links, our data-driven growth model is more rigid and can only perform those growth predictions that are also represented in the image distribution.
Conversely, such models have the advantage that they are not based on model parameters that have uncertainties.
The output of knowledge-based models are parameters, such as the expected yield or biomass. 
Additionally, in the case of simulations, quantitative information about the processes responsible for plant growth can be derived \cite{jame1996cgm}, but not generated images that can be treated like explainable sensor observations and from which any phenotypic traits can be derived.

To obtain a comprehensive image-based growth model for cauliflower on a larger scale, it is necessary to increase the amount of data, including more images of plants of different environmental conditions and time points.
Images are needed that represent as much variability in input factors and accordingly phenotypes as possible, so that the diversity in the distribution of images is large but not biased, which can lead to issues with GANs, as discussed in more detail in the following section on mode collapse.
When increasing the set of images, it is important to ensure that plants in all images have similar basic conditions (e.g., type of soil, climatic zone, season, genome).
Since there is also a natural bias in plant development in dependence of locations, which is not directly reflected in the images, a new independent model should be trained for each region with deviating basic conditions.
In contrast, a model does not need to be limited in time, because the distribution of phenotypes becomes particularly diverse over several years due to different environmental influences.
If the basic conditions are stable, the environmental and management influences have an effect on generated images of the future, as we demonstrated using different treatments for irrigation and fertilization.

Methodologically, growth prediction using cGAN can also be done for other crops than brassica that have different phenotypes, such as wheat or lupines.
Similarly, there is no restriction on the number of plants on an image, so plot- or field-wise image generation and growth prediction are also feasible.
The more plants there are on each image at a correspondingly lower resolution, the more difficult it becomes to evaluate the generated images by single plant detection, so it becomes necessary to evaluate robust field-wise phenotypic traits, such as plant number or total biomass.
Another challenge is the height of many crops, where no ground robot with artificial light source is practical. 
Alternatives are orthophotos, where, however, lighting conditions are different for each pass.
This can be addressed by style transfer methods, also based on GANs \cite{gogoll2020domainAdapt}.
What remains is that with birds-eye view, only a small part of the plant is observed as the height increases, which is why a change of perspective to a side view could be useful.
However, this brings new challenges in the alignment of the images.
The finer the leaf structures and grain ears and the lighter the plant, the less it is ensured anyway that the plants themselves do not move between two points in time, and this issue can be amplified by wind.
While geo-referencing and stable plant positions are crucial requirements for Pix2Pix image alignment, CycleGANs approaches could help for plants where alignment is not achievable \cite{foerster2019,zhu2017unpaired}.

\subsubsection{Mode Collapse}
A common problem when using GANs is mode collapse, which refers to the problem of the model converging to a state where different inputs result in the same or very similar outputs.
The number of modes the generator collapses to varies widely, and often the model jumps back and forth between modes during training iterations.
We observed mode collapse in preliminary experiments with both datasets, in which we learned independent models for each time step.
For instance, in the Brassica dataset, one model only for the time prediction of week 1 $\rightarrow$ 4 and another model only for step 2 $\rightarrow$ 5.
Training one model for all growth stages prevented mode collapse, attributed to higher diversity in the training data. 
However, there are still phenomena that occur similar to mode collapses.
In some cases, two generated plants have the same basic structure, but the plant is more extended in a later growth phase.
That means the inner leaves look the same, while the outer leaves are expanded.
Note that the position of the plant in the generated image remains correct, i.e., even if the inner leaves of the plant appear unchanged, the center position of the generated instance is close to the center position of the input domain.
So the generated plants are still realistic and reasonable.

In order to increase the output diversity, we have experimented with an increase in input diversity by means of data augmentation methods such as Cutout, CutMix \cite{yun2019cutmix} and synthetically generated data.
Moreover, we conducted experiments with changed hyperparameter, modifying $\gamma$ to control the loss weighting (see \figref{eq:pix2pixCmpltObjctv}), or choosing a different architecture such as the Diversity-Sensitive Conditional GAN \cite{yang2019diversity}, which is designed to force variability through a different structure and loss functionality.
None of these attempts is successful in enforcing more diversity in the generated images.
For applications similar to ours, where we primarily evaluate the size of the generated plants, the lack of variability at the leaf level is not a problem. However, this phenomenon shows that the power of the generator is still limited and that reliably increasing diversity is still an open research question.

\subsubsection{Future research}
For future research, we point out some interesting directions:
In modern industrial agriculture, generally, a farmer aims to plan already at the planting stage when the field will need to be cultivated and when crops will be ready for harvest.
However, uncertain long-term weather forecasts, extreme weather events, and pest or pathogen occurrence make it challenging to predict these outcomes with high accuracy.
To account for this, monitoring and screening the plants' current status in the field would be necessary but is labor-intensive and time-consuming if conducted by the farmer or another expert.
As presented in this paper, a monitoring approach, which comprises high-throughput sensor measurements and automatic analysis, can overcome several challenges connected to this.
First, since the current stage of the plant is continuously observed, and the prediction of the future stage is based on it, the estimated time for harvesting is expected to be more accurate than conventional approaches. 
Besides, the difference between a plant's current status and a farmer's expectations about plant status is visually assessible and quantifiable so that the farmer could take early action in the field to prevent negative yield results.
Finally, planning reliability could be increased, not only for the time of harvest but also for the expected harvest yield.

A future methodological direction is to extend the generator model by adding aligned image time series and further environmental input as well as domain knowledge that is available for the field and that influence plant growth.
For instance, for most fields, climate data is available by nearby weather stations that provide data such as temperature, precipitation, and sunshine hours.
There may also be pre-information on soil quality or type and nutrient content, which can be included in the model as bias.
Also, other kinds of image data, such as hyperspectral or thermal images, can be considered.
These inputs can be combined in a multi-modal generative adversarial network,
which is expected to give more comprehensive and reliable growth predictions, resulting in easier planning and enable earlier and more targeted actions in the field.
We further see a high demand for generating a range of possible output images (depending on different input parameters) instead of a single output image, which could support simulations operating within process-based growth models.
Finally, we would like to point out that we have used our methodological approach for crop growth predictions, however, we also see potential in other application areas such environmental sciences, where the prediction of future states is an essential task.

\section{Conclusion}\label{sec:conclusion}
In this work, we have demonstrated the suitability of a conditional generative adversarial network based on Pix2Pix to generate realistic looking and reliably generated images of future plant growth stages in an unsupervised manner. 
In our experiments with laboratory-grown \textit{Arabidopsis~thaliana} and field-grown cauliflower, we comprehensively evaluated the generated images.
The analysis using Fréchet Inception Distance shows quantitatively that the generated images show strong similarities to the reference images.
We qualitatively evaluate our results by applying Mask R-CNN to obtain an instance segmentation of plants. 
In doing so, we derive valuable parameters such as plant size from the estimated instance segmentation mask and use them for comparing plant instances of the generated and real images.
Using our conditional adversarial network-based growth model for cauliflower, we illustrated that the mean plant size and the plant's position are realistically estimated in six different growth stages. 
Analyzing the results for plants with four different field treatments, we demonstrated that plants with good irrigation and fertilization are predicted to be larger than those with deficiencies, as is the case in reality.
Compared to the laboratory experiment with \textit{Arabidopsis~thaliana}, we observed a higher discrepancy between generated and reference images which can be related to the less exact geo-referencing of images and partial overlaps between plants.
We consider this method applicable in agriculture because it adds an explainable component to existing crop growth models by visualizing the phenotype, is sensitive to tiny differences in crop appearance due to different treatments, and is scalable from single crops to field-wise analyses, as well as from brassica to any other crops.

\section*{ACKNOWLEDGEMENTS}\label{ACKNOWLEDGEMENTS}
This work was funded by the Deutsche Forschungsgemeinschaft (DFG, German Research Foundation) under Germany’s Excellence Strategy – EXC 2070 – 390732324. 


\IEEEtriggeratref{49}


\bibliographystyle{IEEEtran}
\bibliography{IEEEabrv,main}
%



\end{document}

%% file: tables/w3_aligned_num.tex

\begin{table}[t]
\centering
\caption{Number of 3-week aligned cauliflower image pairs divided in bed 01 (train) and bed 03 (test). For Experiment \texttt{Cauliflower} all training data of bed 01 are used. The test data are divided into the regions \{\texttt{i+f+}, \texttt{i+f-}, \texttt{i-f+}, \texttt{i-f-}\}. Due to a sensor outage in week 7, there are no image pairs for step 4$\rightarrow$7 in bed 01 and only pairs in section \texttt{i+f-} of bed 03.}
\begin{tabular}{ccccccc}
\toprule
Week           & Train: Bed 01 & \multicolumn{5}{c}{Test: Bed 03}\\
           & $\sum$  & \texttt{i+f+} & \texttt{i+f-} & \texttt{i-f+} & \texttt{i-f-} & $\sum$ \\
\midrule
1 $\rightarrow$ 4 & 124    & 43        & 57         & 44         & 46          & 190   \\ 
2 $\rightarrow$ 5 & 322    & 110       & 82         & 94         & 90          & 376   \\ 
3 $\rightarrow$ 6 & 198    & 72        & 70         & 72         & 80          & 294   \\ 
4 $\rightarrow$ 7 & 0      & 0         & 86         & 0          & 0           & 86    \\ 
5 $\rightarrow$ 8 & 407    & 166       & 100        & 152        & 104         & 522   \\ 
6 $\rightarrow$ 9 & 270    & 148       & 118        & 124        & 150         & 540  \\
\bottomrule
\end{tabular}
\label{tab:w3nums}
\end{table}

%% file: tables/fid_table.tex
\begin{table}[t]
\centering
\caption{Overview about \text{FID} scores \text{FID}($\mathcal{N}_r,\mathcal{N}_g$), \text{FID}($\mathcal{N}_r,\mathcal{N}_t$) and \text{FID}($\mathcal{N}_g,\mathcal{N}_t$) of both different experiments and different treatments.}
\begin{tabular}{lcccc}
\toprule
\multicolumn{2}{l}{Experiment} & \text{FID}($\mathcal{N}_r,\mathcal{N}_g$) & \text{FID}($\mathcal{N}_r,\mathcal{N}_t$) & \text{FID}($\mathcal{N}_g,\mathcal{N}_t$)\\
\midrule
\multirow{4}{*}{\texttt{Cauliflower}}    & \texttt{i-f-} & 34.18           & 55.62           & 51.63           \\
                               & \texttt{i-f+} & 33.91           & 48.04           & 55.29           \\
                               & \texttt{i+f-} & 38.64           & 48.21           & 56.33           \\
                               & \texttt{i+f+} & 31.14           & 49.60           & 54.55           \\
\multicolumn{2}{l}{\texttt{Arabidopsis}}       & 38.12           & 28.90           & 42.25  \\
\bottomrule
\end{tabular}
\label{tab:fid}
\end{table}



%% file: main.bbl
\begin{thebibliography}{10}
\providecommand{\url}[1]{#1}
\csname url@samestyle\endcsname
\providecommand{\newblock}{\relax}
\providecommand{\bibinfo}[2]{#2}
\providecommand{\BIBentrySTDinterwordspacing}{\spaceskip=0pt\relax}
\providecommand{\BIBentryALTinterwordstretchfactor}{4}
\providecommand{\BIBentryALTinterwordspacing}{\spaceskip=\fontdimen2\font plus
\BIBentryALTinterwordstretchfactor\fontdimen3\font minus
  \fontdimen4\font\relax}
\providecommand{\BIBforeignlanguage}[2]{{%
\expandafter\ifx\csname l@#1\endcsname\relax
\typeout{** WARNING: IEEEtran.bst: No hyphenation pattern has been}%
\typeout{** loaded for the language `#1'. Using the pattern for}%
\typeout{** the default language instead.}%
\else
\language=\csname l@#1\endcsname
\fi
#2}}
\providecommand{\BIBdecl}{\relax}
\BIBdecl

\bibitem{fischer2009world}
G.~Fischer, ``World food and agriculture to 2030/50,'' in \emph{Proc.~of the
  FAO Expert Meeting on How to Feed the World in 2050}, 2009, pp. 24--26.

\bibitem{gebbers2010precision}
R.~Gebbers and V.~I. Adamchuk, ``Precision agriculture and food security,''
  \emph{Science}, vol. 327, no. 5967, pp. 828--831, 2010.

\bibitem{kitzes2008shrink}
J.~Kitzes, M.~Wackernagel, J.~Loh, A.~Peller, S.~Goldfinger, D.~Cheng, and
  K.~Tea, ``Shrink and share: humanity's present and future ecological
  footprint,'' \emph{Philosophical Transactions of the Royal Society B:
  Biological Sciences}, vol. 363, no. 1491, pp. 467--475, 2008.

\bibitem{tyagi2016towards}
A.~C. Tyagi, ``Towards a second green revolution,'' \emph{Irrigation and
  Drainage}, vol.~65, no.~4, pp. 388--389, 2016.

\bibitem{kamilaris2018deep}
A.~Kamilaris and F.~X. Prenafeta-Bold{\'u}, ``Deep learning in agriculture: A
  survey,'' \emph{Computers and Electronics in Agriculture}, vol. 147, pp.
  70--90, 2018.

\bibitem{zhu2018deep}
N.~Zhu, X.~Liu, Z.~Liu, K.~Hu, Y.~Wang, J.~Tan, M.~Huang, Q.~Zhu, X.~Ji,
  Y.~Jiang \emph{et~al.}, ``Deep learning for smart agriculture: Concepts,
  tools, applications, and opportunities,'' \emph{International Journal of
  Agricultural and Biological Engineering}, vol.~11, no.~4, pp. 32--44, 2018.

\bibitem{chlingaryan2018machine}
A.~Chlingaryan, S.~Sukkarieh, and B.~Whelan, ``Machine learning approaches for
  crop yield prediction and nitrogen status estimation in precision
  agriculture: A review,'' \emph{Computers and Electronics in Agriculture},
  vol. 151, pp. 61--69, 2018.

\bibitem{kuwata2015estimating}
K.~Kuwata and R.~Shibasaki, ``Estimating crop yields with deep learning and
  remotely sensed data,'' in \emph{Proc.~of the IEEE International Geoscience
  and Remote Sensing Symposium (IGARSS)}, 2015, pp. 858--861.

\bibitem{you2017deep}
J.~You, X.~Li, M.~Low, D.~Lobell, and S.~Ermon, ``Deep gaussian process for
  crop yield prediction based on remote sensing data,'' in \emph{Proc.~of the
  AAAI Conference on Artificial Intelligence}, 2017, pp. 4559--4565.

\bibitem{goodfellow2014generative}
I.~Goodfellow, J.~Pouget-Abadie, M.~Mirza, B.~Xu, D.~Warde-Farley, S.~Ozair,
  A.~Courville, and Y.~Bengio, ``Generative adversarial nets,'' in
  \emph{Proc.~of the Advances in Neural Information Processing Systems
  (NeurIPS)}, 2014, pp. 2672--2680.

\bibitem{ledig2017photo}
C.~Ledig, L.~Theis, F.~Husz{\'a}r, J.~Caballero, A.~Cunningham, A.~Acosta,
  A.~Aitken, A.~Tejani, J.~Totz, Z.~Wang, and W.~Shi, ``Photo-realistic single
  image super-resolution using a generative adversarial network,'' in
  \emph{Proc.~of the IEEE Conference on Computer Vision and Pattern Recognition
  (CVPR)}, 2017, pp. 4681--4690, vgg loss, content loss, style loss.

\bibitem{radford2015unsupervised}
A.~Radford, L.~Metz, and S.~Chintala, ``Unsupervised representation learning
  with deep convolutional generative adversarial networks,'' \emph{arXiv
  preprint arXiv:1511.06434}, 2015.

\bibitem{roscher2020explain}
R.~Roscher, B.~Bohn, M.~Duarte, and J.~Garcke, ``Explain it to me - facing
  remote sensing challenges in the bio- and geosciences with explainable
  machine learning,'' \emph{Proc.~of the ISPRS Annals of Photogrammetry, Remote
  Sensing and Spatial Information Sciences}, vol. V-3-2020, pp. 817--824, 2020.

\bibitem{bender2020high}
A.~Bender, B.~Whelan, and S.~Sukkarieh, ``A high-resolution, multimodal data
  set for agricultural robotics: A ladybird's-eye view of brassica,''
  \emph{Journal of Field Robotics}, vol.~37, no.~1, pp. 73--96, 2020.

\bibitem{isola2017image}
P.~Isola, J.-Y. Zhu, T.~Zhou, and A.~A. Efros, ``Image-to-image translation
  with conditional adversarial networks,'' in \emph{Proc.~of the IEEE
  Conference on Computer Vision and Pattern Recognition (CVPR)}, 2017, pp.
  1125--1134.

\bibitem{heusel2017gans}
M.~Heusel, H.~Ramsauer, T.~Unterthiner, B.~Nessler, and S.~Hochreiter, ``{GAN}s
  trained by a two time-scale update rule converge to a local nash
  equilibrium,'' in \emph{Proc.~of the Advances in Neural Information
  Processing Systems (NeurIPS)}, 2017, pp. 6626--6637.

\bibitem{wu2019detectron2}
Y.~Wu, A.~Kirillov, F.~Massa, W.-Y. Lo, and R.~Girshick, ``Detectron2,''
  \url{https://github.com/facebookresearch/detectron2}, 2019.

\bibitem{feller1995phanologische}
C.~Feller, H.~Bleiholder, L.~Buhr, H.~Hack, M.~Hess, R.~Klose, U.~Meier,
  R.~Stauss, T.~v.~d. Boom, and E.~Weber, ``{P}h{\"a}nologische
  {E}ntwicklungsstadien von {G}em{\"u}sepflanzen {I}. {Z}wiebel-, {W}urzel-,
  {K}nollen- und {B}lattgem{\"u}se,'' \emph{Nachrichtenblatt des Deutschen
  Pflanzenschutzdienstes}, vol.~47, no.~8, pp. 193--205, 1995.

\bibitem{pandey2017impact}
P.~Pandey, V.~Irulappan, M.~V. Bagavathiannan, and M.~Senthil-Kumar, ``Impact
  of combined abiotic and biotic stresses on plant growth and avenues for crop
  improvement by exploiting physio-morphological traits,'' \emph{Frontiers in
  Plant Science}, vol.~8, p. 537, 2017.

\bibitem{kage2001predicting}
H.~Kage, H.~St{\"u}tzel, and C.~Alt, ``Predicting dry matter production of
  cauliflower ({B}rassica oleracea {L}. botrytis) under unstressed conditions:
  Part {II}. comparison of light use efficiency and photosynthesis--respiration
  based modules,'' \emph{Scientia horticulturae}, vol.~87, no.~3, pp. 171--190,
  2001.

\bibitem{miller2001using}
P.~Miller, W.~Lanier, and S.~Brandt, ``Using growing degree days to predict
  plant stages,'' \emph{Montana State University (MT200103 AG 7/2001)}, vol.
  59717, no. 406, pp. 994--2721, 2001.

\bibitem{olesen2000simulation}
J.~E. Olesen and K.~Grevsen, ``A simulation model of climate effects on plant
  productivity and variability in cauliflower ({B}rassica oleracea {L}.
  botrytis),'' \emph{Scientia Horticulturae}, vol.~83, no.~2, pp. 83--107,
  2000.

\bibitem{sihag2019review}
J.~Sihag and D.~Prakash, ``A review: Importance of various modeling techniques
  in agriculture/crop production,'' in \emph{Soft Computing: Theories and
  Applications}.\hskip 1em plus 0.5em minus 0.4em\relax Springer, 2019, pp.
  699--707.

\bibitem{park2005comparison}
S.~Park, C.~Hwang, and P.~Vlek, ``Comparison of adaptive techniques to predict
  crop yield response under varying soil and land management conditions,''
  \emph{Agricultural Systems}, vol.~85, no.~1, pp. 59--81, 2005.

\bibitem{alhnaity2019using}
B.~Alhnaity, S.~Pearson, G.~Leontidis, and S.~Kollias, ``Using deep learning to
  predict plant growth and yield in greenhouse environments,'' \emph{arXiv
  preprint arXiv:1907.00624}, 2019.

\bibitem{johansen2019predicting}
K.~Johansen, M.~Morton, Y.~Malbeteau, B.~Aragon, S.~Almashharawi, M.~Ziliani,
  Y.~Angel, G.~Fiene, S.~Negr{\~a}o, M.~A.~A. Mousa, M.~A. Tester, and M.~F.
  McCabe, ``Predicting biomass and yield at harvest of salt-stressed tomato
  plants using uav imagery,'' \emph{The International Archives of the
  Photogrammetry, Remote Sensing and Spatial Information Sciences}, vol.
  XLII-2/W13, pp. 407--411, 2019.

\bibitem{nevavuori2019crop}
P.~Nevavuori, N.~Narra, and T.~Lipping, ``Crop yield prediction with deep
  convolutional neural networks,'' \emph{Computers and Electronics in
  Agriculture}, vol. 163, p. 104859, 2019.

\bibitem{watt2020phenotyping}
M.~Watt, F.~Fiorani, B.~Usadel, U.~Rascher, O.~Muller, and U.~Schurr,
  ``Phenotyping: New windows into the plant for breeders,'' \emph{Annual Review
  of Plant Biology}, vol.~71, 2020.

\bibitem{foerster2019}
A.~Foerster, J.~Behley, J.~Behmann, and R.~Roscher, ``Hyperspectral plant
  disease forecasting using generative adversarial networks,'' in
  \emph{Proc.~of the IEEE International Geoscience and Remote Sensing Symposium
  (IGARSS)}, 2019, pp. 1793--1796.

\bibitem{li2018unsupervised}
J.~Li, J.~Jia, and D.~Xu, ``Unsupervised representation learning of image-based
  plant disease with deep convolutional generative adversarial networks,'' in
  \emph{Proc.~of the IEEE 37th Chinese Control Conference (CCC)}, 2018, pp.
  9159--9163.

\bibitem{nazki2020unsupervised}
H.~Nazki, S.~Yoon, A.~Fuentes, and D.~S. Park, ``Unsupervised image translation
  using adversarial networks for improved plant disease recognition,''
  \emph{Computers and Electronics in Agriculture}, vol. 168, pp. 105--117,
  2020.

\bibitem{barth2018improved}
R.~Barth, J.~Hemming, and E.~J. van Henten, ``Improved part segmentation
  performance by optimising realism of synthetic images using cycle generative
  adversarial networks,'' \emph{arXiv preprint arXiv:1803.06301}, 2018.

\bibitem{suarez2019image}
P.~L. Su{\'a}rez, A.~D. Sappa, B.~X. Vintimilla, and R.~I. Hammoud, ``Image
  vegetation index through a cycle generative adversarial network,'' in
  \emph{Proc.~of the IEEE/CVF Conference on Computer Vision and Pattern
  Recognition Workshops (CVPRW)}, 2019, pp. 1014--1021.

\bibitem{dai2020crop}
Q.~Dai, X.~Cheng, Y.~Qiao, and Y.~Zhang, ``Crop leaf disease image
  super-resolution and identification with dual attention and topology fusion
  generative adversarial network,'' \emph{IEEE Access}, vol.~8, pp.
  55\,724--55\,735, 2020.

\bibitem{nazki2019image}
H.~Nazki, J.~Lee, S.~Yoon, and D.~S. Park, ``Image-to-image translation with
  gan for synthetic data augmentation in plant disease datasets,'' \emph{Smart
  Media Journal}, vol.~8, no.~2, pp. 46--57, 2019.

\bibitem{zhu2017unpaired}
J.-Y. Zhu, T.~Park, P.~Isola, and A.~A. Efros, ``Unpaired image-to-image
  translation using cycle-consistent adversarial networks,'' in \emph{Proc.~of
  the IEEE International Conference on Computer Vision (ICCV)}, 2017, pp.
  2242--2251.

\bibitem{hong2018conditional}
W.~Hong, Z.~Wang, M.~Yang, and J.~Yuan, ``Conditional generative adversarial
  network for structured domain adaptation,'' in \emph{Proc.~of the IEEE/CVF
  Conference on Computer Vision and Pattern Recognition (CVPR)}, 2018, pp.
  1335--1344.

\bibitem{lin2018conditional}
J.~Lin, Y.~Xia, T.~Qin, Z.~Chen, and T.-Y. Liu, ``Conditional image-to-image
  translation,'' in \emph{Proc.~of the IEEE/CVF Conference on Computer Vision
  and Pattern Recognition (CVPR)}, 2018, pp. 5524--5532.

\bibitem{mirza2014conditional}
M.~Mirza and S.~Osindero, ``Conditional generative adversarial nets,''
  \emph{arXiv preprint arXiv:1411.1784}, 2014.

\bibitem{sun2017revisiting}
C.~Sun, A.~Shrivastava, S.~Singh, and A.~Gupta, ``Revisiting unreasonable
  effectiveness of data in deep learning era,'' in \emph{Proc.~of the IEEE
  International Conference on Computer Vision (ICCV)}, 2017, pp. 843--852.

\bibitem{zhu2018data}
Y.~Zhu, M.~Aoun, M.~Krijn, J.~Vanschoren, and H.~T. Campus, ``Data augmentation
  using conditional generative adversarial networks for leaf counting in
  arabidopsis plants.'' in \emph{Proc.~of the British Machine Vision Conference
  (BMVC)}, 2018, p. 324.

\bibitem{giuffrida2017arigan}
M.~Valerio~Giuffrida, H.~Scharr, and S.~A. Tsaftaris, ``{ARIGAN}: Synthetic
  arabidopsis plants using generative adversarial network,'' in \emph{Proc.~of
  the IEEE International Conference on Computer Vision (ICCV) Workshops}, Oct
  2017, pp. 2064--2071.

\bibitem{antipov2017face}
G.~Antipov, M.~Baccouche, and J.-L. Dugelay, ``Face aging with conditional
  generative adversarial networks,'' in \emph{Proc.~of the IEEE International
  Conference on Image Processing (ICIP)}, 2017, pp. 2089--2093.

\bibitem{borji2019pros}
A.~Borji, ``Pros and cons of gan evaluation measures,'' \emph{Computer Vision
  and Image Understanding}, vol. 179, pp. 41--65, 2019.

\bibitem{arjovsky2017wasserstein}
M.~Arjovsky, S.~Chintala, and L.~Bottou, ``Wasserstein {GAN},'' \emph{arXiv
  preprint arXiv:1701.07875}, 2017.

\bibitem{salimans2016improved}
T.~Salimans, I.~Goodfellow, W.~Zaremba, V.~Cheung, A.~Radford, and X.~Chen,
  ``Improved techniques for training {GAN}s,'' in \emph{Proc.~of the
  International Conference on Neural Information Processing Systems (NeurIPS)},
  2016, pp. 2234--2242.

\bibitem{roscher2014Automated}
R.~Roscher, K.~Herzog, A.~Kunkel, A.~Kicherer, R.~T{\"o}pfer, and
  W.~F{\"o}rstner, ``Automated image analysis framework for high-throughput
  determination of grapevine berry sizes using conditional random fields,''
  \emph{Computers and Electronics in Agriculture}, vol. 100, pp. 148--158,
  2014.

\bibitem{zabawa2020counting}
L.~Zabawa, A.~Kicherer, L.~Klingbeil, R.~T{\"o}pfer, H.~Kuhlmann, and
  R.~Roscher, ``Counting of grapevine berries in images via semantic
  segmentation using convolutional neural networks,'' \emph{ISPRS Journal of
  Photogrammetry and Remote Sensing}, vol. 164, pp. 73--83, 2020.

\bibitem{hamuda2016survey}
E.~Hamuda, M.~Glavin, and E.~Jones, ``A survey of image processing techniques
  for plant extraction and segmentation in the field,'' \emph{Computers and
  Electronics in Agriculture}, vol. 125, pp. 184--199, 2016.

\bibitem{he2017mask}
K.~He, G.~Gkioxari, P.~Doll{\'a}r, and R.~Girshick, ``Mask {R-CNN},'' in
  \emph{Proc.~of the IEEE International Conference on Computer Vision (ICCV)},
  2017, pp. 2961--2969.

\bibitem{bell_jonathan_2016_168158}
J.~Bell and H.~M. Dee, ``Aberystwyth leaf evaluation dataset,'' Nov 2016.

\bibitem{ronneberger2015u}
O.~Ronneberger, P.~Fischer, and T.~Brox, ``{U-Net}: Convolutional networks for
  biomedical image segmentation,'' in \emph{Proc.~of International Conference
  on Medical Image Computing and Computer-Assisted Intervention
  (MICCAI)}.\hskip 1em plus 0.5em minus 0.4em\relax Springer, Cham, 2015, pp.
  234--241.

\bibitem{lin2014microsoft}
T.-Y. Lin, M.~Maire, S.~Belongie, J.~Hays, P.~Perona, D.~Ramanan,
  P.~Doll{\'a}r, and C.~L. Zitnick, ``Microsoft {COCO}: Common objects in
  context,'' in \emph{Proc.~of the European Conference on Computer Vision
  (ECCV)}.\hskip 1em plus 0.5em minus 0.4em\relax Springer, Cham, 2014, pp.
  740--755.

\bibitem{di2016overview}
A.~Di~Paola, R.~Valentini, and M.~Santini, ``An overview of available crop
  growth and yield models for studies and assessments in agriculture,''
  \emph{Journal of the Science of Food and Agriculture}, vol.~96, no.~3, pp.
  709--714, 2016.

\bibitem{jame1996cgm}
Y.~Jame and H.~Cutforth, ``Crop growth models for decision support systems,''
  \emph{Canadian Journal of Plant Science}, vol.~76, 01 1996.

\bibitem{gogoll2020domainAdapt}
D.~Gogoll, P.~Lottes, J.~Weyler, N.~Petrinic, and C.~Stachniss, ``Unsupervised
  domain adaptation for transferring plant classification systems to new field
  environments, crops, and robots,'' in \emph{Proc.~of the IEEE/RSJ
  International Conference on Intelligent Robots and Systems (IROS)}, 2020, pp.
  2636--2642.

\bibitem{yun2019cutmix}
S.~Yun, D.~Han, S.~J. Oh, S.~Chun, J.~Choe, and Y.~Yoo, ``Cutmix:
  Regularization strategy to train strong classifiers with localizable
  features,'' in \emph{Proc.~of the IEEE International Conference on Computer
  Vision (ICCV)}, 2019, pp. 6023--6032.

\bibitem{yang2019diversity}
D.~Yang, S.~Hong, Y.~Jang, T.~Zhao, and H.~Lee, ``Diversity-sensitive
  conditional generative adversarial networks,'' \emph{arXiv preprint
  arXiv:1901.09024}, 2019.

\end{thebibliography}
